\documentclass[pdflatex,sn-mathphys-num]{sn-jnl}

\usepackage{graphicx}%
\usepackage{multirow}%
\usepackage{amsmath,amssymb,amsfonts,amsthm}%
\usepackage{mathrsfs}%
\usepackage[title]{appendix}%
\usepackage{xcolor}%
\usepackage{textcomp}%
\usepackage{manyfoot}%
\usepackage{booktabs}%
\usepackage{makecell}%
\usepackage{algorithm}%
\usepackage{algorithmicx}%
\usepackage{algpseudocode}%
\usepackage{listings}%
\usepackage{tikz}
\usetikzlibrary{positioning, calc, arrows.meta, fit, backgrounds}

\definecolor{slate50}{HTML}{F8FAFC}
\definecolor{slate300}{HTML}{CBD5E1}
\definecolor{slate600}{HTML}{475569}
\definecolor{slate800}{HTML}{0F172A}

\definecolor{amber50}{HTML}{FFFBEB}
\definecolor{amber700}{HTML}{B45309}

\definecolor{emerald50}{HTML}{ECFDF5}
\definecolor{emerald700}{HTML}{047857}

\definecolor{rose50}{HTML}{FEF2F2}
\definecolor{rose700}{HTML}{B91C1C}

\usepackage{bm}
\usepackage{microtype}
\usepackage{enumitem}
 
\newtheorem{proposition}{Proposition}
\newcommand{\YES}{\mathrm{ACCEPT}}
\newcommand{\NO}{\mathrm{REJECT}}
\newcommand{\ind}{\mathbf{1}}
\newcommand{\Prob}{\mathbb{P}}

\usepackage{xr-hyper}    
\usepackage{hyperref}    
\usepackage{zref-base}

\makeatletter
\newcommand*{\addFileDependency}[1]{%
  \typeout{(#1)}
  \@addtofilelist{#1}
  \IfFileExists{#1}{}{\typeout{No file #1.}}
}
\makeatother

\theoremstyle{thmstyleone}%

\theoremstyle{thmstyletwo}%

\theoremstyle{thmstylethree}%

\begin{document}

\title[Certified Compression for Clinical LLMs]{Do We Need the Largest LLMs for Every Clinical Decision?}

\author*[1,2]{\fnm{Ali H.} \sur{Lazem}}\email{lhl23prg@bangor.ac.uk}
\author[1]{\fnm{William} \sur{Teahan}}\email{w.j.teahan@bangor.ac.uk}

\affil*[1]{\orgdiv{School of Computer Science and Engineering},
\orgname{Bangor University}, \orgaddress{\city{Bangor},
\postcode{LL57 2DG}, \state{Gwynedd}, \country{United Kingdom}}}
\affil[2]{\orgname{University of Thi-Qar}, \orgaddress{\city{Nasiriyah},
\postcode{64001}, \country{Iraq}}}


\abstract{Large clinical language models often require multi-GPU infrastructure, limiting local
deployment. We introduce CERTA, a decision-level statistical certificate that determines when a
quantized model may substitute for its native BF16 reference and escalates uncertifiable decisions to it.
Across eleven instruction-tuned models, both general-purpose and domain-adapted, from three families (7--72B), we
evaluated 2{,}500 clinical-verification decisions from real de-identified patient contexts
(2{,}484 patients), with patient-disjoint calibration and testing. Eight-bit round-to-nearest
quantization, evaluated at BF16 storage, certified 99.8--100\% of decisions for every model in the higher-power operational analysis. In the same
analysis, the class-conditional four-bit frontier reduced the amortized, BF16-equivalent deployment cost to 25.8\% and 27.3\% of the BF16 reference for Qwen2.5-72B and
HuatuoGPT-o1-72B respectively, with measured peak memory of 43.0 and 42.7\,GB, below the 48\,GB
capacity of a single accelerator (a measured allocation below the device's nominal capacity, not an end-to-end deployment on it); under the exact split-sample certificate, which additionally accounts for configuration selection, these flagship costs were 26.7\% and 27.9\%. On external disease named-entity benchmarks against human annotations, the frontier-selected NF4 configurations had a mean absolute F1 change of 0.045 on both datasets, with substantial model-specific variation. CERTA converts compression from a fixed precision choice into a
risk-controlled deployment policy for this verification task.
}




\maketitle

Large language models are increasingly capable on clinical text~\cite{singhal2023clinical,thirunavukarasu2023medicine}, yet the most capable open-weight
models are also among the largest, and running a 70-billion-parameter model in BF16 (bfloat16), its native precision,
requires accelerator memory that can exceed the capacity available in many healthcare and research settings. A single such
model in its native precision occupies well over one hundred gigabytes and spans two high-end
datacenter GPUs. This puts a practical barrier between the models that perform best on clinical
language and the settings where clinical data is most sensibly kept, at the point of care, inside an
institution's firewall, on a modest local server.
 
Quantization is the usual route to a smaller model. Lowering the numerical precision of the weights
shrinks the memory footprint, sometimes several-fold, and various literature reports that aggregate
accuracy stays close to the original after quantization~\cite{frantar2023gptq,dettmers2022llmint8,dettmers2023qlora,lin2024awq}. Aggregate accuracy, however, is the wrong
currency for a clinical decision. Two models can match on a benchmark score while disagreeing on
individual cases, and in a verification or triage step, it is the individual decision---accept this
finding or reject it---that carries the consequence. A method that preserves an average can still
flip the specific calls that matter, and standard benchmarks offer no per-decision guarantee that
the smaller model may stand in for the larger one.
 
Our question is therefore narrower and more operational than whether the compressed model is still
accurate: for the decision in front of us, can the compressed model be used in place of the
BF16 reference model without changing the decision, and can that substitution be controlled at a
stated error rate? Posed this way, the choice of precision stops being a fixed global setting and
becomes a per-decision, per-model question with a testable answer.
 
\begin{figure}[t]
\centering
\resizebox{\textwidth}{!}{
\begin{tikzpicture}[
  font=\sffamily,
  >=Latex,
  card/.style={
    draw=slate300, fill=white, rounded corners=3pt, align=center,
    text width=54mm, inner sep=6pt, font=\small},
  certaBox/.style={
    draw=amber700, fill=amber50, rounded corners=3pt, align=center,
    text width=54mm, inner sep=6pt, line width=0.9pt, font=\small},
  useBox/.style={
    draw=emerald700, fill=emerald50, rounded corners=3pt, align=center,
    text width=42mm, inner sep=5pt, font=\small, line width=0.7pt},
  escBox/.style={
    draw=rose700, fill=rose50, rounded corners=3pt, align=center,
    text width=42mm, inner sep=5pt, font=\small, line width=0.7pt},
  subCard/.style={
    draw=slate300, fill=slate50, rounded corners=3pt, align=center,
    text width=42mm, inner sep=5pt, font=\small},
  vramBadge/.style={
    draw=emerald700, fill=emerald50, rounded corners=3pt, align=center,
    text width=30mm, inner sep=4pt, font=\small\bfseries, text=emerald700},
  arr/.style={-{Latex[length=2.4mm]}, line width=0.8pt, color=slate600},
  line/.style={line width=0.8pt, color=slate600},
  lbl/.style={font=\scriptsize\sffamily, text=slate600, fill=white, inner sep=2pt},
  brlbl/.style={font=\scriptsize\sffamily\bfseries, fill=white, inner sep=2pt}
]
  \node[card, text width=40mm, fill=slate50] (input)
    {\textbf{Clinical input}\\[0.5mm]\textcolor{slate600}{$x$}};
  \node[card, below=10mm of input] (quant) {\textbf{QUANTIZED MODEL}\\[1mm]
    \textcolor{slate600}{$Q_k$, 4 / 6 / 8 bit}};
  \draw[arr] (input) -- (quant);
  \node[certaBox, below=12mm of quant] (certa) {\textbf{CERTA ROUTER}\\[1mm]
    threshold for the predicted class\\[0.3mm] vs.\ target risk level $\alpha$};
  \draw[arr] (quant) -- node[lbl] {decision $Q_k(x)$,\ margin $m_k(x)$} (certa);
  \node[subCard, right=12mm of quant, text width=34mm, draw=slate300, dash pattern=on 2pt off 2pt] (offline)
    {\footnotesize $Q_k$ is an offline quantization\\ of the BF16 reference $F$};
  \draw[-{Latex[length=2mm]}, dash pattern=on 2pt off 2pt, color=slate600] (offline) -- (quant);
  \coordinate (fork)      at ($(certa.south) + (0,-7mm)$);
  \coordinate (forkLeft)  at ($(fork) + (-28mm,0)$);
  \coordinate (forkRight) at ($(fork) + ( 28mm,0)$);
  \draw[line] (certa.south) -- (fork);
  \draw[line] (forkLeft) -- (forkRight);
  \node[useBox, below=13mm of forkLeft] (use) {\textbf{RETURN QUANTIZED $Q_k(x)$}\\[0.3mm]
    \textcolor{slate600}{\footnotesize no BF16 call for this certified input}};
  \draw[arr] (forkLeft) -- node[brlbl, text=emerald700, pos=0.42] {CERTIFIED} (use);
  \node[escBox, below=13mm of forkRight, text width=46mm] (esc)
    {\textbf{INVOKE BF16 REFERENCE $F$}\\[0.3mm]
    \textcolor{slate600}{\footnotesize fallback only; $\sim$142--146 GB}};
  \draw[arr] (forkRight) -- node[brlbl, text=rose700, pos=0.42] {UNCERTIFIED} (esc);
  \node[subCard, below=8mm of esc]  (fullref) {full-reference decision $F(x)$};
  \draw[arr] (esc) -- (fullref);
  \node[vramBadge, below=8mm of use] (vramEnd) {measured peak $\sim$41--43 GB\\[0.3mm]
    \textcolor{slate600}{\footnotesize\mdseries below nominal 48 GB (untested)}};
  \draw[arr] (use) -- (vramEnd);
\end{tikzpicture}
}
\caption{\textbf{CERTA turns compression into a per-decision certify-or-escalate policy.}
For each clinical input $x$, the quantized candidate $Q_k$ (at $4$, $6$ or $8$ bits) proposes a
decision, and CERTA compares its decision margin $m_k(x)$ against a threshold calibrated to a risk
level $\alpha$. Decisions that clear the certificate are served by $Q_k$ on the low-memory path;
the rest are escalated to the BF16 reference $F$. The $70$--$72$B reference occupies
$\sim$$142$--$146$\,GB and spans two $80$\,GB accelerators, whereas the certified NF4 path has a measured
peak of $\sim$$41$--$43$\,GB, below the nominal capacity of a single $48$\,GB accelerator (deployment on
that specific hardware was not directly tested). The reference must remain reachable whenever the policy
permits escalation. For a clinical accept/reject decision,
the threshold is calibrated separately per decision class (class-conditional, Mondrian).}
\label{fig:architecture}
\end{figure}
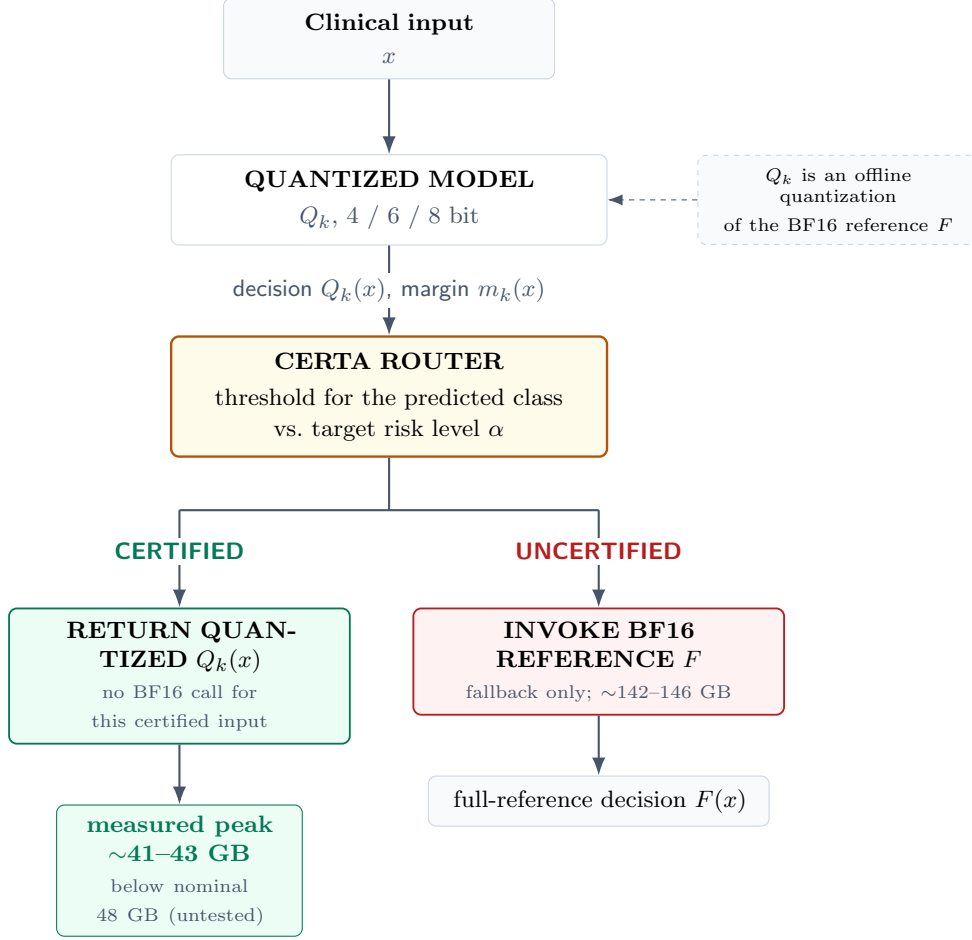

We answer it with CERTA, a procedure that certifies or escalates each decision (Fig.~\ref{fig:architecture}). Using a calibration
set in which quantized decisions are labelled by their agreement with the BF16 reference, CERTA learns a
finite-sample threshold~\cite{angelopoulos2025learn,angelopoulos2023gentle,angelopoulos2024crc} on the quantized model's decision margin that bounds, at a chosen risk level, the rate at
which a served decision disagrees with the reference. At inference the router uses only the quantized
model's decision, margin and this calibrated threshold: decisions that clear it are served by the
compressed model, and the rest are escalated to the reference, which is invoked only on escalation. Repeating this across
bit-widths and quantizers gives, for each model, the least costly configuration the certificate will
accept at a given risk, which we call its certified compression frontier. Because a clinical
accept/reject decision has two sides with different consequences, the deployable policy controls the
reject decision separately from the accept decision. The decisions we study are a deliberately narrow but
clinically relevant primitive, the binary verification of candidate entities and answers extracted from
clinical narratives, rather than diagnosis, treatment selection, triage or prognosis; the framework is
general, but the evidence reported here concerns this verification step.
 
Across eleven models from three families, spanning 7 to 72 billion parameters and including both
general-purpose and clinically adapted checkpoints, evaluated on 2{,}500 verifier decisions constructed from real de-identified patient contexts
drawn from 2{,}484 patients, eight-bit round-to-nearest quantization at BF16 storage was certified for every model, and a four-bit
scheme reduced the amortized, BF16-equivalent deployment cost to about a quarter of the BF16 size for the strongest
models, bringing a 72-billion-parameter model's measured footprint from two large accelerators within the capacity of one. The question in our
title, whether the largest model is needed for every clinical decision, has a concrete answer: for
most decisions, a compressed model suffices and can be certified to do so, and the certificate
identifies the minority that must be referred to the full model.

\section*{Results}
 
We evaluated CERTA on the accept/reject decision of an in-house clinical verifier, using 2{,}500
decisions drawn from 2{,}484 patients in the PMC-Patients corpus and split patient-disjointly into 1{,}250 for calibration and
1{,}250 for held-out test, across eleven models, two quantizers and risk levels from $1\%$ to
$10\%$.
 
\subsection*{Eight-bit substitution is certified for every model}
 
At $8$-bit precision and a $5\%$ target risk ($\alpha=5\%$), the certificate accepted the quantized model on essentially every decision (the certify-or-escalate policy is shown in Fig.~\ref{fig:architecture}). Certified
coverage was $99.8\%$ to $100\%$ for all eleven models, with held-out empirical certified-set agreement of
$97.8\%$ to $99.5\%$ (Fig.~\ref{fig:landscape}a; per-configuration
certified coverage for every model, at $\alpha=5\%$, is in Supplementary Table~3). This universal
$8$-bit result came from round-to-nearest quantization, which we evaluated at BF16 storage as a
numerical-precision reference; it therefore establishes that these decisions are robust to $8$-bit rounding,
not that $8$-bit halves the deployed memory. The memory-reducing $8$-bit route, \texttt{LLM.int8()}, is a separate
implementation that certified the same decisions for most models but failed on two BF16-native $70$B
checkpoints (see below), so numerical $8$-bit robustness and deployable $8$-bit memory reduction are distinct properties. The result was uniform
across general-purpose and clinically adapted checkpoints, across the Llama, Qwen and Mistral
families, and across the $7$B to $72$B range, and the earlier fragility we had observed on a
smaller synthetic set did not survive the move to real prompts. Across configurations, certified coverage fell well below raw decision preservation, because the certificate rejects low-margin cases (Fig.~\ref{fig:landscape}b). At $6$-bit the certificate remained
strong for nine of the eleven models, each with at least $98\%$ certified coverage (for example
MMed-Llama-3-8B at $98.2\%$); the two exceptions were MMed-Llama-3.1-70B at $93.6\%$ and
BioMistral-7B at $80.3\%$.
 
Round-to-nearest quantization at $4$-bit, by contrast, was strongly model-dependent. Certified coverage ranged
from $0\%$ for MMed-Llama-3-8B to $91$--$94\%$ for HuatuoGPT-o1-72B and Qwen2.5-72B, with the
medical fine-tunes of the shared Llama-3 backbone falling across the whole interval. This split, an
$8$-bit result that holds for every model against a $4$-bit result that holds for only some, is the
problem the rest of the study addresses: a fixed precision is either too conservative for the robust
models or fails the prespecified decision-risk criterion for the fragile ones.
 
\subsection*{The certified compression frontier is model and quantizer specific}
 
A genuinely memory-reducing $4$-bit quantizer changed the picture. NF4 raised raw $4$-bit decision preservation for all
eleven models relative to round-to-nearest, and its certified coverage at $4$-bit ranged from
$82.2\%$ to $100\%$. Qwen2.5-72B and HuatuoGPT-o1-72B certified at $100\%$, and the domain
fine-tunes that round-to-nearest had failed now certified at high coverage under NF4, OpenBioLLM-70B
at $98.1\%$, Med42-70B at $97.5\%$, MMed-Llama-3.1-70B at $96.9\%$ and BioMistral-7B at $95.8\%$ (Fig.~\ref{fig:quantizer}).

Taking the lower envelope over the deployable quantizers at a $5\%$ target risk (denoted $\alpha$), the certified compression
frontier selected NF4 for all eleven models and settled between $25.0\%$ and $38.4\%$ of the
BF16 size. Qwen2.5-72B and HuatuoGPT-o1-72B reached the $4$-bit floor at $25.0\%$, a fourfold
reduction, with no escalation. Llama-3.3-70B sat highest, at $38.4\%$, because its NF4 certified coverage
($82.2\%$) is lower, so more of its decisions escalate to the reference. NF4 was preferred over $8$-bit
\texttt{LLM.int8()} wherever it certified, following a break-even structure in which moving
from $k$ to $k'$ bits reduces amortized size only once certified coverage exceeds
$(16-k)/(16-k')$; this accounts for the pattern of bit-width selection and for why the deployable frontier
rests at $4$-bit. Round-to-nearest, which retains BF16 storage, served as the numerical-precision
baseline rather than a deployment option, so the two quantizer families were complementary in role, and
the certificate, not a prior preference for one scheme, set the deployed configuration.
 
\subsection*{A risk level sets the minimum deployable footprint}
 
Sweeping the risk level turned the frontier into a per-model risk--resource curve. At a stringent
$1\%$ target risk, every one of the eleven models still had a certified operating point,
between $31.2\%$ and $62.8\%$ of BF16 size. At $5\%$ the range was $25.0\%$ to $38.4\%$, and at
$10\%$ it tightened to $25.0\%$ to $30.2\%$, with eight of the eleven models at the $4$-bit floor. The
curves were monotonic and stable across the evaluated risk levels, so an operator can read the deployable footprint directly against
whatever disagreement risk a setting permits (Fig.~\ref{fig:risk}a; the frontiers for all eleven
models appear in Supplementary Fig.~1, and are robust to the confidence level,
calibration size, threshold-grid resolution and the choice of calibration/test split, Supplementary
Figs~5 and~6 and Supplementary Tables~9, 10, 11, and 12).
 
The certificate was well calibrated in aggregate. At each model's deployed operating point, the
disagreement observed on the held-out test set stayed at or below the target risk for every model and
quantizer across $\alpha$ from $1\%$ to $10\%$ (at most $0.6\%$ at $\alpha=1\%$ and $3.9\%$ at
$\alpha=5\%$); the closest approach to a target was Med42-70B under round-to-nearest $4$-bit, at $9.9\%$
against the $10\%$ target (Fig.~\ref{fig:risk}b).
 
\subsection*{Class-conditional certification protects the reject decision}
 
Aggregate calibration, however, does not imply calibration within a decision class, and the
distinction matters clinically because a reject decision suppresses an unsupported output.
Partitioning by the quantized model's own decision, the marginal certificate violated the per-class
$5\%$ target risk for six model-classes at NF4 $4$-bit, four of them on the reject side. Among
predicted $\NO$ decisions the observed disagreement reached $17.4\%$ for OpenBioLLM-70B, $10.0\%$ for
MMed-Llama-3-8B, $6.1\%$ for MMed-Llama-3.1-70B and $6.1\%$ for Mistral-7B; the accept-side violations
were Llama-3-70B at $6.9\%$ and Qwen2.5-72B at $5.2\%$. These failures were not specific to NF4:
at round-to-nearest $8$-bit the reject class of MMed-Llama-3.1-70B still exceeded the target risk. A high
aggregate coverage concealed these.
 
Class-conditional (Mondrian) certification removed them (Fig.~\ref{fig:risk}c). Under this operational
class-conditional analysis, with a separate threshold per predicted class, each decision class either had
an admissible upper bound at or below the target risk or was escalated in full (the formal finite-sample
guarantee is provided by the exact split-sample procedure):
OpenBioLLM-70B, MMed-Llama-3.1-70B and BioMistral-7B escalated their entire reject class rather than certify it
unsafely, for two distinct reasons: the reject classes of OpenBioLLM-70B and BioMistral-7B are too small
to certify at the target risk, whereas MMed-Llama-3.1-70B has adequate calibration resolution but too
many reject-side disagreements for any threshold to clear the target (Supplementary
Table~6 and Supplementary Fig.~4). The remaining reject classes were certified at reduced coverage under a stricter
per-class threshold (MMed-Llama-3-8B, for instance, at $35.1\%$ reject-class coverage), while the
accept class continued to certify at high coverage (MMed-Llama-3.1-70B at $96.5\%$). In the higher-power
operational analysis of this clinically appropriate policy, the class-conditional frontier at
$\alpha=5\%$ spanned $25.8\%$ to $40.2\%$ of BF16-equivalent nominal-bit cost (Table~\ref{tab:frontier});
because its threshold grid is built on the same calibration data used for the bounds, these are
operating-point estimates rather than the formal certificate. The formal finite-sample guarantee at
confidence $1-\delta$ ($\delta=0.10$) is supplied by the exact split-sample procedure, which independently
constructs the threshold grid, uses one decision per patient in certification, and corrects for selection
across both decision classes and the two deployable configurations; under it the frontier spans $26.7\%$
to $48.4\%$, with all eleven models still selecting NF4 (Supplementary Table~7). The stricter class-conditional policy
selected NF4 for all eleven models, and at $\alpha=5\%$ the compressed model served most decisions,
escalating only $1.1\%$ (Qwen2.5-72B) to $20.2\%$ (MMed-Llama-3-8B) of them to the reference. The flagship survived the
stricter guarantee: Qwen2.5-72B and HuatuoGPT-o1-72B deployed at $25.8\%$ and $27.3\%$ at
$\alpha=5\%$, and reached the exact fourfold point ($25.0\%$) at a modest risk level, $\alpha\ge 8\%$
for Qwen and $\alpha\ge 9\%$ for HuatuoGPT. Across models the class-conditional guarantee changed amortized size by $-0.8$ to $+7.9$
percentage points relative to the marginal certificate (median $+2.3$), the largest increases
falling on the models that escalate their entire reject class, in exchange for controlling risk
on both sides of the decision. The selected per-class thresholds and Clopper--Pearson~\cite{clopper1934} bounds for
every frontier point are in Supplementary Table~5, and the decision-risk audit in
Supplementary Table~4; the class-conditional frontier is itself stable across the
confidence and risk levels (Supplementary Tables~9 and~13).
 
\subsection*{Certified compression changes the hardware footprint of 70--72B models}
 
The frontier translated into a measured change in the accelerators a model requires (Fig.~\ref{fig:landscape}c). Peak memory for
the $70$B and $72$B models fell from $141.7$ to $146.3$\,GB at BF16, which spans two $80$\,GB
devices, to a measured $41.0$ to $43.0$\,GB under NF4, below the nominal capacity of a single $48$\,GB device; $8$-bit \texttt{LLM.int8()}
sat between them at $73.4$ to $77.6$\,GB. These measurements indicate that the compressed models fit within
that capacity, but deployment on a specific $48$\,GB device was not directly tested. The $4$-bit floor of $25\%$ is a weight-memory ratio; the
measured peak-VRAM ratio is higher (for example $43.0/146.3\approx29\%$ for Qwen2.5-72B) because of
quantization metadata and runtime allocations. At operating points with zero observed escalation, no BF16
fallback was invoked on the evaluated decisions, and the measured peak allocation is below the nominal
$48$\,GB capacity of a single device. This does not establish that the reference can be removed from a
general-purpose deployment: zero observed escalations in a finite sample do not imply that no future input
will escalate, so the reference must remain reachable whenever the routing policy permits escalation. The memory reduction carried a throughput cost in the kernel used: the $70$B
round-to-nearest configurations ran at $8.6$ to $9.1$ prompts per second against roughly $5.0$ for
NF4, so we report the two together and make no joint claim of smaller and faster. Separately, the
memory-reducing $8$-bit route, \texttt{LLM.int8()}, produced degenerate outputs on two BF16-native $70$B
checkpoints, Llama-3-70B and Med42-70B (its float16 activation casting fails on these models); running the
certification pipeline on the recorded outputs assigned them $0\%$ deployable coverage. This is an
engineering failure of that $8$-bit implementation, not a statistical certification failure on a
well-formed quantized model, and it illustrates that a nominal $8$-bit route is not automatically a
deployable one. Because round-to-nearest retains BF16 storage,
there is no universally deployable $8$-bit point, and the certified memory reduction is realised at $4$-bit through NF4.
 
\subsection*{Decision preservation and external task accuracy are complementary}
 
Finally, we asked whether the certified configurations preserve task accuracy against human labels,
a question the substitution certificate does not answer. On the two disease named-entity benchmarks (Fig.~\ref{fig:external}),
$8$- and $6$-bit round-to-nearest quantization changed the F1 score negligibly, with mean absolute changes of $0.005$ and
$0.019$ on BC5CDR-Disease and $0.011$ and $0.018$ on NCBI-Disease; $4$-bit round-to-nearest quantization degraded
it further (mean absolute change $0.057$ and $0.044$). Critically, we benchmarked the exact NF4
$4$-bit configurations that the frontier deploys, and their effect on the F1 score was comparable to
round-to-nearest quantization's, and smaller for several models: a mean absolute change of $0.045$ on BC5CDR-Disease and $0.045$ on NCBI-Disease, with Qwen2.5-72B
essentially unchanged ($-0.008$ and $0.000$). The clearest illustration is MMed-Llama-3.1-70B, the
worst round-to-nearest $4$-bit casualty ($-0.163$ and $-0.198$), which improved under the deployed
NF4 configuration ($+0.078$ and $+0.091$); the apparent collapse of domain fine-tunes at $4$-bit was
largely an artifact of round-to-nearest rather than a property of the configuration CERTA actually
selects. Model-specific drops remained (OpenBioLLM-70B $-0.132$ on BC5CDR, Med42-70B $-0.111$ on
NCBI, HuatuoGPT-o1-72B $-0.107$ on BC5CDR), so task performance was largely, not perfectly,
preserved (precision, recall and F1 in Supplementary Table~16; a paired per-document
human-gold correctness analysis in Supplementary Tables~17 and~18;
the two verifier tasks certified consistently, Supplementary Table~15).
 
Reference agreement and task accuracy were related but not interchangeable. Across models the
per-model association between certified coverage and F1 retention was weak, and the eleven
configurations occupied distinct positions in this plane,
which is expected because the certificate concerns agreement with the BF16 reference on the
verifier task, whereas the benchmark concerns correctness against human annotations on a different
task. The two therefore provide complementary evidence: the certificate controls what compression
changes relative to the reference, and the benchmark shows that the frontier-selected configurations also
hold up against human labels.

\section*{Discussion}
 
CERTA reframes compression as a controlled substitution problem rather than a fixed precision
choice. Instead of quantizing a clinical model to a chosen bit-width and trusting a benchmark
average, we certify, for each decision, that the compressed model agrees with its BF16
reference at a stated risk, and we escalate the decisions that cannot be certified. This makes the
deployable precision a property of the model and the risk level rather than a global default, and
the certificate does real work: certified coverage was systematically lower than raw agreement,
because the procedure rejects the low-margin cases that raw accuracy would wave through. Sweeping
bit-widths and two quantizers turns the certificate into a per-model frontier that selects, at each
risk level, the least costly configuration that meets the guarantee, and the frontier chose a real
4-bit quantizer (NF4) for all eleven models rather than any single scheme fixed in advance. Two
comparisons make the practical value concrete. Uniform 8-bit rounding is numerically certified across
all eleven models, but that arm retains BF16 storage and is not a half-memory deployment; the
memory-reducing 8-bit route, \texttt{LLM.int8()}, certified most models yet failed on two 70B checkpoints. CERTA
instead reaches about a quarter of BF16-equivalent nominal-bit cost for the strongest models by certifying
the harder 4-bit regime only where the guarantee holds; the measured peak VRAM reduction is smaller in
percentage terms (for example, Qwen2.5-72B falls from $146.3$ to $43.0$\,GB, a measured $29.4\%$), because
physical memory also includes quantization metadata and runtime allocations. A threshold based only on
uncalibrated confidence would not carry the same finite-sample risk guarantee. Certified coverage ran
systematically below raw decision preservation, by about fifteen percentage points on average, which
illustrates that many decisions that happen to agree with the reference remain uncertified under the
chosen risk constraint; raw agreement with the reference is also not observable at deployment without
running the reference itself.
 
The resulting change in hardware is concrete. Peak memory for the 70- and 72-billion-parameter
models fell from roughly 142 to 146 gigabytes, which spans two 80-gigabyte accelerators, to a measured 41
to 43 gigabytes, below the nominal capacity of a single 48-gigabyte device (deployment on that specific
hardware was not directly tested); where a model certified with no observed escalation, no reference call
was triggered on the evaluated decisions, though the reference must stay reachable for any deployment that
permits escalation. Reducing the measured inference footprint of these
models to the capacity of a single 48-gigabyte accelerator could lower the infrastructure barrier to running
them locally, where clinical data can remain inside an institution, subject to that institution's
security, governance and data-protection controls. This benefit is specifically in
memory: the low-precision kernel we used carried a throughput cost, so we report memory and speed
together and do not claim a joint improvement. Whether specialized low-precision kernels narrow this
throughput gap is an engineering question outside the scope of the certificate.
 
CERTA bounds
disagreement with the BF16 reference; it does not bound the clinical correctness of that
reference, and a preserved decision that was wrong in BF16 remains wrong. We therefore
treated correctness as a separate question and evaluated it against human annotations on independent
disease named-entity benchmarks, where the deployed 4-bit configurations preserved task performance;
tellingly, a domain model that appeared to collapse under one 4-bit quantizer recovered under the
one CERTA actually deploys. The association between certified agreement and this external accuracy
was weak, which is expected because they answer different questions, and we present them as
complementary evidence rather than as one validating the other. The same discipline drove the
class-conditional result. A single threshold controls disagreement only on average, and for models
whose decisions skew toward acceptance it can leave the rarer reject decision, whose error
consequences may differ by workflow, poorly controlled; certifying each decision class separately restores control on both sides at a
small additional cost, and we take this class-conditional policy, not the marginal one, as the
appropriate default for a clinical accept/reject decision.
 
Several boundaries define the scope of these findings. The guarantee is a finite-sample, high-probability
control under patient-level iid sampling (and hence exchangeability between calibration and deployment),
not a deterministic proof for any single patient, and it is specific to the task on
which the certificate was calibrated, so a deployment would certify on its own decision distribution
rather than extrapolate from ours. The memory reduction depends on the quantization scheme and its
kernel, and escalation requires the reference model to be reachable. At operating points with zero
observed escalation, the BF16 reference was not invoked for any evaluated decision; this does not
establish that future inputs will never be escalated, so a deployment that permits escalation must keep
the reference reachable. Our evaluation balanced
accept and reject cases as a deliberate stress test; a prevalence-shift stress test (Supplementary
Fig.~2 and Supplementary Table~14) showed that the marginal
certificate can exceed the target risk when the deployment accept/reject mix departs from
calibration, whereas the class-conditional policy stays controlled across the range, with $95\%$ bootstrap confidence
intervals reported for the coverage and F1 estimates in the corresponding Supplementary analyses. The decisions themselves, however, were certified
against the BF16 reference model, not adjudicated by clinicians. The most valuable next step is precisely that adjudication: a study in which clinicians
grade the severity of the cases where the compressed and BF16 reference models disagree would connect
compression to clinical consequence, and would move the evidence from same decision as the full
model to clinically inconsequential when it differs. CERTA identifies the low-margin decisions that are
enriched for compression-induced disagreement, namely the ones it escalates, and can therefore prioritise
them for clinician adjudication; comparing the recorded quantized and BF16 outputs then isolates the
subset on which the models actually disagree. The routing margin alone does not identify exactly which
inputs will disagree: some escalated inputs agree with the reference and some certified inputs can still
differ, but the escalated set has systematically lower margin and higher disagreement
(Supplementary Fig.~3), which makes such a study directly actionable. The certificate also admits any prespecified routing statistic:
it uses a scalar decision margin as the service score, but the same certify-or-escalate guarantee holds for
any score, including a learned substitution-risk score, provided the score and threshold grid are fixed
independently of the certification sample (for example, learned on a separate selection subset). Such a
score could widen the certifiable region without changing the statistical framework.

\section*{Methods}

\subsection*{Study design and estimand}
 
CERTA was designed around a single deployment question: can a quantized clinical language model
be substituted for the BF16 reference model it is intended to replace while controlling
compression-induced changes to individual decisions at a prespecified rate? The \emph{primary
estimand} is the probability that the quantized and BF16 reference models return different
decisions among the inputs on which the quantized model is permitted to act without escalation.
The clinical correctness of the reference model is a \emph{separate estimand}; it is neither
estimated nor bounded by the certificate, and it is studied only through the independent
human-annotated benchmarks described below. Keeping these two estimands apart is a deliberate
design choice, because a certificate that preserved the reference decision would still inherit any
error the reference itself makes.
 
The study has three stages. The first establishes a decision-level statistical certificate that
decides, from the quantized model's output alone, whether substitution is defensible at risk
$\alpha$. The second turns the certificate into a certify-or-escalate deployment policy and, by
sweeping bit-widths, quantizers and risk levels, into a model-specific compression frontier. The
third evaluates the certified configurations on external tasks with human labels, which measure
correctness rather than agreement. All certification analyses used real clinical-verifier
decisions sampled from a de-identified patient corpus, with calibration and test partitions
separated at the level of the patient.
 
\subsection*{Models}
 
We evaluated eleven autoregressive, instruction-tuned language-model checkpoints, general-purpose and domain-adapted, spanning three architecture
families (Llama, Qwen, Mistral), parameter scales from 7B to 72B, and both general-purpose and
clinically adapted checkpoints: Llama-3-8B, Llama-3-70B and Llama-3.3-70B~\cite{grattafiori2024llama3}, Qwen2.5-72B~\cite{qwen2024qwen25} and
Mistral-7B~\cite{jiang2023mistral}; and Med42-70B~\cite{christophe2024med42v2}, OpenBioLLM-70B~\cite{pal2024openbiollm}, MMed-Llama-3-8B and MMed-Llama-3.1-70B~\cite{qiu2024mmedlm}, BioMistral-7B~\cite{labrak2024biomistral} and
the reasoning-oriented HuatuoGPT-o1-72B~\cite{chen2024huatuogpto1} (Supplementary Table~19 lists the eleven checkpoints with their families, adaptation type and parameter counts). The set was chosen to separate three properties that are
usually confounded in compression studies, parameter count, architecture family and clinical
adaptation, rather than to identify a single best-performing clinical model; the general-purpose and domain-adapted
checkpoints within the Llama family share closely related Llama-3 backbones so that the effect of the adaptation recipe can be
read against a largely fixed baseline. All checkpoints were loaded locally with the Hugging Face Transformers library~\cite{wolf2020transformers} and PyTorch~\cite{paszke2019pytorch}, with no network access during
evaluation, and the native BF16 realization of each checkpoint served as the reference. Because these
checkpoints are released in this 16-bit format, BF16 is their native precision, and all compression
results are reported relative to this BF16 reference.
 
\subsection*{Clinical decision task and prompt construction}
 
The certified decision was the accept/reject step of an in-house clinical verification
pipeline that we have not yet published. In this pipeline, a generator model proposes candidate clinical
entities and answers extracted from de-identified patient narratives, and a verifier model judges
each candidate, accepting supported outputs and rejecting unsupported ones; this binary verification
step is the single decision CERTA certifies, and no other component of the pipeline is used or
evaluated here. The decision
space was binary: $\YES$ and $\NO$. We evaluated two candidate types. In
named-entity verification the model judged whether a candidate entity, presented with a proposed
category, was a valid and specific clinical entity of that category. In answer verification the
model judged whether a proposed answer was supported by a supplied context and question. Both used
the pipeline's verbatim prompt templates and constrained the model to a single-token accept/reject
answer. Prompts were rendered with each model's native chat template; for checkpoints whose
template does not accept a system role, the system instruction was merged into the user turn.
 
Candidates were sampled in equal numbers from the two task types and balanced across accept and
reject. Reject cases were constructed by replacing an answer with a mismatched answer drawn from
the candidate pool, or by replacing a named-entity category with a different clinical category, so
that the evaluation contains constructed reject cases rather than an accept-only distribution; these
substitutions were not individually verified as unsupported, so an occasional constructed reject may
remain plausible in context. Patient
assignment to the calibration or test partition (below) was fixed before candidate corruption and
sampling, so no information from a patient crosses the calibration/test boundary. The balanced
design is a deliberate stress test of decision-level certification; the reference and quantized
decision distributions per model and task are given in Supplementary Table~2, and a
prevalence-shift analysis in Supplementary Fig.~2.
 
\subsection*{Patient corpus and evaluation set}
 
Candidates were drawn from PMC-Patients~\cite{zhao2023large}, a publicly available dataset of
167{,}034 de-identified patient summaries curated from open-access case reports in PubMed Central.
Because the summaries are already de-identified and publicly released, no direct identifiers
were accessible during the study. The evaluation set contained 2{,}500 decisions, 1{,}250 from
named-entity verification and 1{,}250 from answer verification, originating from 2{,}484 distinct
patient summaries; sixteen patients contributed a second decision. Each candidate retained its source
patient identifier from PMC-Patients, which was used to enforce the patient-disjoint split.
 
\subsection*{BF16 reference and compression-induced disagreement}
 
For input $x$, the BF16 reference produced answer-token logits $z_F(x,\YES)$ and
$z_F(x,\NO)$, and the $k$-bit quantized model produced $z_k(x,\YES)$ and $z_k(x,\NO)$. The decision
of a model is the higher-scoring answer token,
\begin{equation}
\begin{aligned}
Q_k(x) &=
\begin{cases}
\YES, & z_k(x,\YES) > z_k(x,\NO),\\
\NO, & \text{otherwise,}
\end{cases} \\[6pt]
F(x) &=
\begin{cases}
\YES, & z_F(x,\YES) > z_F(x,\NO),\\
\NO, & \text{otherwise.}
\end{cases}
\end{aligned}
\end{equation}
and the quantized decision margin is
\begin{equation}
m_k(x)=\bigl|\,z_k(x,\YES)-z_k(x,\NO)\,\bigr|.
\end{equation}
The certificate reads the \emph{quantized} model's margin $m_k$, which is available at inference,
and never the reference margin. The compression-induced disagreement event is
\begin{equation}
D_k(x)=\ind\{\,Q_k(x)\neq F(x)\,\}.
\end{equation}
CERTA controls the rate of $D_k$; it does not estimate the reference error
$\ind\{F(x)\neq y\}$ against any clinical label $y$.
 
For every model and every quantization configuration, the reference and quantized models were run
once over all 2{,}500 inputs, and we stored, per input, the reference decision $F(x)$, the
quantized decision $Q_k(x)$ and the quantized margin $m_k(x)$. These recorded quantities are the
only inputs to the certification, routing, frontier and class-conditional procedures below; no
further model execution is required to compute any reported certificate.
 
\subsection*{Decision-level statistical certification}
 
A certificate is a rule of the form $m_k(x)\ge\tau$ (Supplementary Table~1 defines the notation used throughout). Writing $A_\tau=\{x:m_k(x)\ge\tau\}$ for the
certified region, the certified-set disagreement risk is the conditional probability
\begin{equation}
R_k(\tau)=\Prob\!\left(\,Q_k(X)\neq F(X)\ \middle|\ m_k(X)\ge\tau\,\right).
\end{equation}
Given a target risk $\alpha$ and confidence parameter $\delta$, CERTA selects $\tau$ so that
$R_k(\tau)\le\alpha$ holds with confidence at least $1-\delta$ under patient-level iid sampling, by the Learn-then-Test procedure~\cite{angelopoulos2025learn} on
a calibration set that is disjoint from the test set.
 
Let $\mathcal{T}=\{t_1,\dots,t_G\}$ be a grid of $G$ candidate thresholds, taken as the empirical
quantiles of the calibration margins. For each $t\in\mathcal{T}$ let $n(t)$ be the number of
calibration inputs with $m_k\ge t$ and $x(t)$ the number of those that disagree with the reference. The
reported certificate uses the full calibration set for both the grid and the bounds, a higher-power
estimate; an exact finite-sample guarantee, in which the grid is built on an independent
threshold-selection subset and the bounds evaluated on a disjoint certification subset, is given by the
split-sample variant (Supplementary Information, Algorithm~S1 and Proposition~S1).
Let $\overline{R}(t)$ be the one-sided Clopper--Pearson~\cite{clopper1934,brown2001interval} upper confidence limit for a binomial
proportion with $x(t)$ successes in $n(t)$ trials at confidence level $1-\delta/G$, computed through
the regularized incomplete beta inverse,
\begin{equation}
\overline{R}(t)=
\begin{cases}
1, & n(t)=0\ \text{or}\ x(t)=n(t),\\[2pt]
\mathrm{Beta}^{-1}_{\,1-\delta/G}\!\left(x(t)+1,\ n(t)-x(t)\right), & \text{otherwise,}
\end{cases}
\end{equation}
where the boundary convention assigns the vacuous upper bound $\overline{R}(t)=1$ when no decisions are
served ($n(t)=0$) or all served decisions disagree ($x(t)=n(t)$), matching the certification algorithm
(Algorithm~S1). The beta-inverse form is exact and numerically stable at the
calibration sizes used here, whereas a direct binomial-coefficient evaluation overflows double precision
for $n(t)$ beyond roughly one thousand.
A threshold is admissible if $\overline{R}(t)\le\alpha$, and CERTA selects the smallest admissible
threshold,
\begin{equation}
\tau=\min\{\,t\in\mathcal{T}: \overline{R}(t)\le\alpha\,\},
\end{equation}
which maximizes certified coverage subject to the risk bound. If no candidate threshold is admissible,
CERTA sets $\tau=\infty$ and escalates every input to the BF16 reference, so the configuration serves no
compressed decisions (the case reported as zero certified coverage); the risk statement is then vacuous
because no input is served by the compressed model. The factor $\delta/G$ is a Bonferroni~\cite{dunn1961bonferroni}
correction that makes the bound hold simultaneously across the grid; the exact finite-sample guarantee
for the split-sample variant, whose grid is independent of the certification sample, is stated in the
proposition below (Proposition~\ref{prop:main}; proved as Supplementary Proposition~S1).
 
\begin{proposition}[Finite-sample certified risk control]\label{prop:main}
Fix a model, quantizer and bit-width $k$, and let the calibration and deployment decisions be
independent and identically distributed at the patient level (exchangeable across the calibration and
test partitions)~\cite{vovk2005algorithmic}.
For the split-sample certificate, whose candidate grid $\mathcal{T}$ is fixed independently of the
certification sample and whose threshold $\tau^\star$ is selected by the rule above on that sample,
\begin{equation*}
\Prob\!\left(\tau^\star<\infty\ \Longrightarrow\ R_k(\tau^\star)\le\alpha\right)\ \ge\ 1-\delta
\end{equation*}
over the calibration draw; that is, whenever a finite threshold is selected, a future exchangeable input
with $m_k(x)\ge\tau^\star$ disagrees with the reference with conditional probability at most $\alpha$. If
no candidate threshold is admissible, $\tau^\star=\infty$: the entire class is escalated, no input is
served by the quantized model, and the conditional risk $R_k(\tau^\star)$ is undefined on the empty served
set. Complete escalation is therefore not a certified zero-risk quantized policy; it serves no compressed
decisions.
\end{proposition}
 
This is a population-level conditional-risk statement over future exchangeable inputs that satisfy
the certification rule. It is not a deterministic guarantee for any individual patient, and we do
not claim per-input certainty. The split-sample certificate to which it applies handles every source of
data-dependent selection: its candidate grid is built on the
$250$-decision selection subset, an outer Bonferroni
over the deployable configurations covers the choice of
quantizer and bit-width, and its Clopper--Pearson bounds are evaluated on the $1{,}000$-decision certification
subset reduced to one decision per patient ($996$ decisions); selection, certification and the $1{,}250$-decision
test set are mutually patient-disjoint, so the count is exactly binomial at the patient level (Supplementary Information,
Algorithm~S1 and Proposition~S1). The frontier reported in the main text is a higher-power operational
analysis that instead uses the full calibration set for both the grid and the bounds; because its grid is
therefore data-dependent, the formal finite-sample guarantee is the split-sample certificate above, and
the two correspond closely (median difference $1.0$, maximum $8.2$ percentage points of BF16 size across the
eleven models, with every model still selecting NF4 $4$-bit; Supplementary Table~7). All
configuration and bit-width choices are made on the calibration data alone; the held-out test set is
used only once, for final evaluation of the already-selected operating point, and never to select a
model, quantizer, threshold or risk level. Throughout we used $G=50$ quantile thresholds,
$\delta=0.10$, and we report the target risk $\alpha$ on a grid from $1\%$ to $10\%$ in one-percent
steps. Each $\alpha$ is a prespecified deployment policy that yields its own finite-sample guarantee;
the frontier over $\alpha$ is a sensitivity analysis, not a licence to select $\alpha$ after observing
test outcomes.
 
\subsection*{Certify-or-escalate deployment policy}
 
The certificate defines an operational router that serves certified inputs with the compressed
model and escalates the rest to the reference,
\begin{equation}
\pi_{k,\tau}(x)=
\begin{cases}
Q_k(x), & m_k(x)\ge\tau,\\
F(x), & m_k(x)<\tau.
\end{cases}
\end{equation}
Certified coverage is the proportion served by the compressed model, $C=\Prob(m_k(X)\ge\tau)$, and
the escalation rate is $E=1-C$. The amortized precision, the coverage-weighted cost in bits that
prices the fallback, simplifies to
\begin{equation}
B_{\mathrm{amortised}}(k,\tau)=100\,\frac{kC+16\,(1-C)}{16}=100\left[1-\left(1-\tfrac{k}{16}\right)C\right]\quad(\%\ \text{of BF16}),
\end{equation}
where $k$ is the nominal bit-width of the compressed configuration and $C$ its certified coverage.
Amortized size is a BF16-equivalent nominal-bit-width cost, not a measurement of physical memory: it
excludes quantization metadata, activations and implementation overhead, and measured peak VRAM is
reported separately (Fig.~\ref{fig:landscape}c and Supplementary Table~19). For
example, Qwen2.5-72B's NF4 configuration has a measured peak of $43.0$\,GB against $146.3$\,GB at BF16
($29.4\%$), somewhat larger than its nominal amortized size. We report certified coverage and the
\emph{empirical certified-set agreement} $\widehat A=\Prob(Q_k(X)=F(X)\mid m_k(X)\ge\tau)$ measured
on held-out data, each with $95\%$ bootstrap confidence intervals. We treat $\widehat A$ as an
out-of-sample diagnostic; the formal guarantee on $R_k(\tau)$ is the one supplied by
Learn-then-Test, and it is not established by the observed test-set agreement.
 
For inputs the router certifies, the reference model need not be resident. For inputs it escalates,
the reference is reached through whatever fallback the deployment configures, a second accelerator,
host offload or a remote endpoint. An operating point with zero \emph{observed} escalation is one in which
the reference was not invoked on the evaluated decisions; because a finite sample cannot establish that no
future input will escalate, a deployment that permits escalation must keep the reference reachable, so its
cost is amortized rather than eliminated.
 
\subsection*{Certified compression frontier}
 
For each model and risk level $\alpha$, we evaluated the certified operating points of the deployable,
memory-reducing configurations (NF4 $4$-bit and $8$-bit \texttt{LLM.int8()}) and selected the one of minimum
amortized size; round-to-nearest, which retains BF16 storage, is a numerical-precision baseline and
is not eligible for the deployment frontier. The certified compression frontier is the lower envelope
over these deployable configurations. The true risk $R_{k,q}\le\alpha$ is the population-level objective,
but it is not observed; the procedure selects configurations using the certificate's Clopper--Pearson
upper bound $\overline{R}_{k,q}$,
\begin{equation}
(k^\star,q^\star)=\operatorname*{arg\,min}_{k,q}\ B_{\mathrm{amortised}}(k,q)
\quad\text{subject to}\quad \overline{R}_{k,q}\le\alpha,
\end{equation}
where for the class-conditional frontier the admissibility condition $\overline{R}_{k,q}\le\alpha$ must
hold separately for each served class and a class with no admissible threshold is escalated in full, so
that the certificate, rather than a prior choice of scheme, selects the deployed configuration.
Configuration selection uses calibration-side information only: a configuration is deployable at $\alpha$
only if its threshold is admissible on the calibration data, and among deployable configurations we take
the one of minimum amortized size computed from calibration-side coverage. In practice this selects NF4
$4$-bit for every model: its calibration-side certified coverage exceeds the break-even level at which its
amortized nominal-bit cost falls below that of $8$-bit \texttt{LLM.int8()}, namely
$C_{\mathrm{NF4}}>\tfrac{2}{3}\,C_{\mathrm{INT8}}$ (which holds for all eleven models, and holds trivially
for the two models where \texttt{LLM.int8()} does not certify), so the choice does not depend on the test
set. The held-out test set is then used only to estimate the coverage
and amortized cost of the selected operating point, so the reported frontier sizes are out-of-sample
estimates for configurations chosen independently of the test outcomes. This configuration selection is
itself data-dependent, so the exact finite-sample certificate (Proposition~\ref{prop:main}; Supplementary
Remark~S1) applies an outer Bonferroni correction across the $H$ deployable configurations.
For a fixed configuration, the marginal certificate allocates $\delta/G$ across its threshold grid, while
the class-conditional certificate allocates $\delta/(2G)$ across the two classes and their grids. For the
exact frontier, which additionally selects among $H=2$ deployable configurations, these become $\delta/(GH)$
and $\delta/(2GH)$ respectively; with $\delta=0.10$, $G=50$ and $H=2$, the exact class-conditional
allocation is $\delta/(2GH)=5\times10^{-4}$. The higher-power operational frontier reported here is a higher-power
operating-point estimate rather than the formal certificate, for two reasons: it builds the threshold grid
and evaluates the Clopper--Pearson bounds on the same calibration decisions, and it uses all calibration
decisions rather than one per patient (sixteen patients contribute two). It omits the outer
configuration-selection factor and reproduces the exactly corrected frontier within a median of $1.0$
percentage point (Supplementary Table~7). The finite-sample guarantee is stated for each
model and risk level, for the threshold-, class- and configuration-selection problem described; it is not
a single family-wise guarantee across all eleven models and all risk levels simultaneously. Sweeping
$\alpha$ from $1\%$ to $10\%$ yields a model-specific risk--resource curve.
 
The bit-width along the frontier obeys a break-even condition. For a fully certified $8$-bit configuration
and an NF4 $4$-bit configuration serving a fraction $c$ of inputs while escalating the rest at $16$ bits,
the $4$-bit configuration has the lower amortized nominal-bit cost when $c>(16-8)/(16-4)=2/3$. More
generally, a $k'$-bit configuration with certified coverage $c$ is cheaper than a fully certified $k$-bit
configuration exactly when
\begin{equation}
c>\frac{16-k}{16-k'}.
\end{equation}
This cost identity motivates the NF4-versus-\texttt{LLM.int8()} comparison that drives the frontier; actual
selection is restricted to the memory-reducing configurations evaluated here (NF4 $4$-bit and $8$-bit
\texttt{LLM.int8()}), which is why the frontier settles at NF4 $4$-bit wherever it certifies.
 
\subsection*{Class-conditional (Mondrian) certification}
 
A single threshold controls $R_k(\tau)$ marginally, averaged over the whole input population, and
does not in general control the disagreement rate within a decision class. For a model whose
decisions are skewed toward one answer this is consequential: when the model rarely returns $\NO$,
the marginal certificate is dominated by the majority $\YES$ decisions and can certify the rare
$\NO$ decisions even when the quantized model preserves them poorly. Because $\NO$ is the reject
decision, this is the direction in which a silent failure is least acceptable clinically. We
therefore treat CERTA as a framework that controls risk at the granularity the deployment requires,
and adopt class-conditional (Mondrian) certification~\cite{vovk2003mondrian} as the default policy for the clinical
accept/reject decision.
 
Classes are defined by the quantized model's own predicted decision $\widehat{g}(x)=Q_k(x)\in\{\YES,\NO\}$
(we write $\widehat{g}$ for the predicted class to avoid clashing with the grid size $G$),
which is observable at inference and therefore usable for routing before the reference decision is
known. For each class $g$ we calibrate a class-specific threshold $\tau_g$ by the same
Learn-then-Test procedure applied to the calibration inputs with $Q_k(x)=g$. The class-conditional
disagreement risk is the population quantity
\begin{equation}
R_g(\tau_g):=\Prob\!\left(\,Q_k(X)\neq F(X)\ \middle|\ m_k(X)\ge\tau_g,\ Q_k(X)=g\,\right),
\end{equation}
which the certificate does not observe directly; the procedure selects $\tau_g$ using the Clopper--Pearson
upper bound and controls $R_g(\tau_g)$ with high probability, as stated below.
Because two class-conditional certificates are calibrated jointly, each class receives a Bonferroni
allocation of $\delta/(2G)$ across its threshold grid, rather than $\delta/G$, so that both decision
classes are controlled simultaneously,
\begin{equation}
\Prob\!\left(\bigcap_{g\in\{\YES,\NO\}}\left\{\tau_g^\star<\infty\ \Longrightarrow\ R_g(\tau_g^\star)\le\alpha\right\}\right)\ge 1-\delta,
\end{equation}
over the calibration draw. Thus every class served by the compressed model satisfies the target risk
bound, while a fully escalated class ($\tau_g^\star=\infty$) serves no compressed decisions and
contributes no compression-induced disagreement to the routed deployment decision. This formal statement
applies to the exact split-sample procedure with the corresponding confidence allocation (Supplementary
Proposition~S1); the full-calibration results are higher-power operating-point estimates. The deployed policy is
\begin{equation}
\pi(x)=
\begin{cases}
Q_k(x), & m_k(x)\ge\tau_{Q_k(x)},\\
F(x), & \text{otherwise,}
\end{cases}
\end{equation}
so that for each predicted decision class CERTA either applies a class-specific threshold whose
finite-sample disagreement bound is at most $\alpha$, or routes that class's uncertifiable inputs
to the reference. The marginal certificate is retained as the simpler baseline. No additional model
inference is required, because the predicted class and the quantized margin are already available
from the quantized forward pass.
 
\subsection*{Quantization methods}
 
Two independent quantization routes were evaluated so that the frontier reflects a property of the
certificate rather than of any single implementation. Round-to-nearest quantization used symmetric
per-group scales over groups of $128$ output-channel elements, with the per-group scale selected by
a mean-squared-error search over clipping ratios $\{0.7,0.8,0.9,1.0\}$, applied in place to each
two-dimensional weight matrix. Round-to-nearest was evaluated at arbitrary bit-width and retained
BF16 storage, providing a controlled precision sweep and an implementation-independent 8-bit
comparison. The second route used the \texttt{bitsandbytes} library: 4-bit NF4 with double quantization~\cite{dettmers2023qlora} and
BF16 compute, and 8-bit \texttt{LLM.int8()}~\cite{dettmers2022llmint8}. The NF4 and \texttt{LLM.int8()} configurations reduce resident
memory and were used for the deployment measurements.
 
Both routes were evaluated on equal footing, and the frontier selects among the configurations that
satisfy the certificate. The \texttt{LLM.int8()} path casts activations to float16 and, for two
BF16-native 70B checkpoints (Llama-3-70B and Med42-70B), produces degenerate outputs; this is an
implementation (engineering) failure of that $8$-bit route on these models. Running the certification
pipeline on the recorded degenerate outputs assigns them zero deployable coverage, which the pipeline
handles gracefully, but these are infeasible configurations rather than well-formed quantized models that
the certificate rejected; a nominal $8$-bit route is therefore not automatically a deployable one.
 
\subsection*{Patient-disjoint calibration and test design}
 
For each model and configuration, the $2{,}500$ decisions were partitioned into $1{,}250$ for
calibration and $1{,}250$ for held-out test. Patient identifiers were grouped before splitting, and
whole patients were allocated to calibration until the target size was reached, with the remainder
allocated to test, so that all decisions from a patient lie on one side of the partition. We
verified for every reported split that no patient identifier occurred in both the calibration and
test sets. All marginal, frontier and class-conditional results use this patient-disjoint partition.
Because sixteen of the $2{,}484$ patients contribute a second decision, we additionally re-ran the
entire certificate with one decision per patient, which reproduces every frontier to within
$1.8$ percentage points of BF16 size and leaves the flagship operating points essentially unchanged
(Supplementary Information).
 
\subsection*{Statistical analysis}
 
Risk levels were evaluated at $\alpha\in\{0.01,0.02,\dots,0.10\}$ with $\delta=0.10$ and a
threshold grid of size $G=50$. The certificate itself is supplied by the Learn-then-Test procedure
with the Clopper--Pearson bound and the Bonferroni correction, stated in full as Supplementary
Algorithm~S1, with its finite-sample guarantee in
Supplementary Proposition~S1; the bootstrap~\cite{efron1979bootstrap} is used only to place
$95\%$ confidence intervals around observed operating-point statistics (certified coverage,
empirical certified-set agreement, and external benchmark metrics), from $1{,}000$ resamples. For the
clinical-verifier test set we resample patients as clusters, not individual decisions, respecting the
sixteen patients that contribute two decisions each; for the external named-entity benchmarks, whose
evaluation units are documents, we resample documents. These intervals quantify empirical-estimate
uncertainty; the formal disagreement-risk control is provided by the Clopper--Pearson certificate, not by
the bootstrap. In the higher-power marginal operational analysis, the smallest risk that can be certified
with zero observed disagreements is bounded below by $1-(\delta/G)^{1/n_{\mathrm{cal}}}$, which at
$n_{\mathrm{cal}}=1{,}250$ is approximately $0.5\%$; this is why a $1\%$ target is statistically reachable
at this calibration size. This is not the resolution floor for the exact class-conditional frontier, which
uses the class-specific certification count and the more conservative allocation $\delta/(2GH)$. As an
empirical calibration diagnostic, at each model's deployed operating point and each $\alpha$ we
compared the target risk with the disagreement observed on the held-out test set and report the
maximum exceedance; this diagnostic checks consistency with the guarantee and does not replace it.
Confidence-level ($\delta$), calibration-size, threshold-grid and split-resampling sensitivity
analyses are reported in Supplementary Tables~9, 10, 11
and~12 and Supplementary Figs~5 and~6. All certificate
and bootstrap computations were implemented in Python with NumPy~\cite{harris2020numpy} and
SciPy~\cite{virtanen2020scipy}, the one-sided Clopper--Pearson bound evaluated as the regularized
incomplete beta inverse.
 
\subsection*{Independent external task validation}
 
To measure task performance independently of reference-model agreement, we evaluated all models at
BF16 and under the quantized configurations, including the NF4 4-bit configurations that
define the deployment frontier, on two human-annotated disease named-entity benchmarks,
BC5CDR-Disease~\cite{li2016bc5cdr} and NCBI-Disease~\cite{dogan2014ncbi}, using $200$ held-out BC5CDR-Disease documents and the full $100$-document NCBI-Disease test split. Models generated
disease-entity lists under a disease-focused instruction; the first well-formed list was parsed and
scored against the gold spans at the entity level by overlap match, after whitespace and case
normalization. We report precision, recall and F1, and the change in F1 relative to the
BF16 reference, per model, dataset and precision. This benchmark addresses correctness
against human labels, a different question from decision substitution, and the two are treated as
complementary rather than interchangeable.
 
\subsection*{Deterministic bound}
 
We evaluated whether a worst-case deterministic propagation of the quantization-induced logit
perturbation could yield a useful bound on the change in decision margin. An illustrative worst-case
amplification calculation shows that generic layerwise propagation through a deep transformer can produce
an uninformative bound. Because we did not establish model-specific bounds on the initial perturbation or
on each layer's operator norm, this calculation is not a numerical bound on the final logit perturbation
for the evaluated models; it documents why this generic deterministic approach did not provide a useful
certificate at the evaluated scale, which is why the certificate is built on the finite-sample,
distribution-level argument above. The calculation is given in the Supplementary Information.
 
\subsection*{Compute and memory measurement}
 
All inference measurements were made on NVIDIA H200 accelerators under the SLURM scheduler~\cite{yoo2003slurm}. A 70B or
72B BF16 reference occupies both accelerators of a two-GPU allocation, and the NF4 4-bit
configurations of these models have a measured peak allocation below the 48\,GB capacity of a single device (we did not execute on that specific hardware). Peak VRAM was recorded as the
maximum memory allocated during the forward pass over the evaluation set, summed over the devices
required by each configuration, and includes model weights together with runtime allocations,
measured with an identical procedure across configurations. Throughput was measured at batch size
one as prompts processed per second. Memory and throughput are reported together, because the
memory-reduced routes carry a dequantization overhead in the kernel used and we make no joint
improvement claim.
 
\section*{Code availability}
 
The CERTA reference implementation is openly available at \url{https://github.com/Ali-Lazem/CERTA}. It provides the complete
finite-sample certification procedure, comprising Learn-then-Test calibration with one-sided
Clopper--Pearson upper bounds and a Bonferroni correction across the threshold grid, the class-conditional
(Mondrian) extension with joint risk control across decision classes, the threshold-calibration routine,
and the certify-or-escalate serving policy. It also includes the full analysis and figure-generation
pipeline, which regenerates every CERTA certificate, table and figure in the main text and the
Supplementary Information directly from the recorded per-decision data, with no model inference required;
the external named-entity benchmarks additionally draw on the released benchmark outputs and scoring code
together with the public gold corpora, as detailed under Data availability. The analysis dependencies are
specified in an \texttt{environment.yml} and a \texttt{requirements.txt}, and the released version is
tagged at the commit used to produce the reported results. An archived snapshot of the code and the
per-decision records is deposited at Zenodo (DOI:
\href{https://doi.org/10.5281/zenodo.23435475}{10.5281/zenodo.23435475}).
 
The calibrated CERTA artifacts are released as a curated Hugging Face collection at
\url{https://huggingface.co/collections/lazem/certa-certified-compression-for-clinical-decisions-6ad373efa6e143550bc15848}.
The collection comprises 53 repositories: 33 round-to-nearest variants (8-, 6- and 4-bit for each of the
eleven models) as numerical-precision evaluation variants, 11 NF4 4-bit variants, and 9 \texttt{LLM.int8()}
8-bit variants for deployment. The two \texttt{LLM.int8()} configurations that failed to produce usable
inference outputs (Llama-3-70B and Med42-70B; the \texttt{LLM.int8()} path casts activations to float16 and
yields degenerate outputs for these BF16-native checkpoints) are not released as checkpoints. Their zero
certified coverage is recorded in the certificate metadata and in the audit tables, marking them as
engineering infeasibility (a failed inference configuration) rather than a statistical certification
failure on a successful quantized run. Each
repository contains the CERTA certification layer for its model, that is, the calibrated marginal and
class-conditional decision thresholds at every evaluated risk level, the machine-readable certificate with
its finite-sample guarantee, and a self-contained serving handler that applies the certify-or-escalate
policy and loads the corresponding checkpoint. The collection includes the round-to-nearest evaluation
variants, which retain BF16 storage and are numerical-precision baselines rather than memory-reducing
deployment variants, alongside the memory-reducing NF4 and \texttt{LLM.int8()} variants; serving
configurations intended for memory-reduced deployment use the corresponding deployable quantized
checkpoints. The two quantization routes evaluated in the study,
round-to-nearest and the \texttt{bitsandbytes} NF4 and \texttt{LLM.int8()}
implementations~\cite{dettmers2023qlora,dettmers2022llmint8}, are included. To respect upstream licensing,
CERTA does not re-host the original base-model weights. Each quantized checkpoint is released only where
the applicable upstream licence permits redistribution of modified weights, under the terms of that
upstream licence; where a licence restricts redistribution or gates access, the checkpoint is not
released and only the CERTA layer and instructions to reproduce the quantization are provided. Each
released repository ships the applicable base-model licence or licences, an Apache-2.0 licence for the
CERTA-authored components, and an explicit record of the base model and its provenance. The precise
redistribution terms are those of each upstream licence rather than a single uniform classification.
 
\section*{Data availability}
 
The clinical accept/reject decisions analysed in this study were derived from de-identified patient
summaries in the publicly available PMC-Patients corpus~\cite{zhao2023large}. External task validation
used the publicly available BC5CDR-Disease~\cite{li2016bc5cdr} and NCBI-Disease~\cite{dogan2014ncbi}
disease named-entity recognition benchmarks. The complete per-decision records underlying every reported
certification result, comprising the reference-model decision, the quantized-model decision and the
decision margin for each input and configuration, are released with the code above and regenerate every
CERTA certificate, table and figure in the main text and Supplementary Information without further model
inference. The external named-entity F1 results (Fig.~\ref{fig:external} and Supplementary
Table~16) are reproduced from the recorded benchmark outputs (per-model, per-dataset
precision, recall and F1) released with the code, scored by the included overlap scorer against the
public BC5CDR-Disease and NCBI-Disease gold corpora; the per-entity paired analysis
(Supplementary Tables~17 and~18), and regeneration of the benchmark
outputs themselves, require re-running the released inference and scoring code on those public benchmark
documents with the model weights, as documented in the code. The base-model checkpoints are publicly
available from their original repositories and are listed, with their exact identifiers, in Supplementary
Table~19. The in-house clinical verification pipeline that produced the accept/reject
decisions is the subject of a separate forthcoming study and is not released here; the released
per-decision records are nonetheless sufficient to reproduce every CERTA certification result reported in
this work.

\section*{Author contributions}
A.H.L. conceived the study, performed the analysis, and wrote the
manuscript. W.T. supervised the work and revised the manuscript.

\section*{Acknowledgements}
We acknowledge the services of Supercomputing Wales and the Bangor
eResearch Team, including the support provided by Dr. Ade Fewings.

\section*{Funding}
This work was supported by the Ministry of Higher Education and Scientific
Research in Iraq. No grant number was assigned.

\section*{Competing interests}
The authors declare no competing interests.

\section*{Ethics}
This study is a retrospective, computational analysis of PMC-Patients~\cite{zhao2023large}, a publicly
available dataset of de-identified patient summaries curated from open-access case reports in PubMed
Central. It involved no patient contact, no change to patient care, and no access to direct identifiers
at any stage. Under the applicable institutional policy, secondary analysis of this kind, using only
publicly available, de-identified records, did not require human-subjects or ethics approval or informed
consent.

\begin{table*}[t]
\centering

\caption{\textbf{Class-conditional certified deployment frontier at
$\alpha=5\%$ (higher-power operational analysis).}
For each model, the configuration is selected using calibration-side information; class-conditional
coverage, overall coverage and amortized size are then estimated on the held-out, patient-disjoint test
set, so the costs shown are out-of-sample estimates for a test-independent selection. The table reports the lowest-amortized-cost configuration
satisfying the class-conditional (Mondrian) CERTA risk constraint within
each predicted decision class, with uncertifiable inputs routed to the
BF16 reference. In the evaluated candidate set, all eleven
frontier points select NF4 at 4 bits. Cov$_{\mathrm{A}}$ and Cov$_{\mathrm{R}}$
denote certified coverage within the predicted ACCEPT and REJECT
classes on the held-out, patient-disjoint test set. An italicized zero means
that the corresponding predicted class is fully escalated because it cannot
be certified at the target risk. Overall coverage and escalation are the
test-set fractions served by the quantized model and routed to the reference,
respectively. Amortized size is the coverage-weighted deployment cost relative
to the 16-bit BF16 reference. BF16 and NF4 are measured peak VRAM. For the
seven 70--72B models, the NF4 compressed model has a measured peak below the nominal capacity of a single
48-GB accelerator (not directly tested on that hardware). The reference must remain reachable wherever the
operating point permits escalation; zero observed escalation on the evaluated decisions does not remove
this requirement for unseen inputs. Qwen2.5-72B and
HuatuoGPT-o1-72B reach the $4$-bit floor of 25\% BF16-equivalent size under the marginal frontier at
$\alpha\ge4\%$ and $\alpha\ge5\%$, and under the stricter class-conditional frontier at $\alpha\ge8\%$ and
$\alpha\ge9\%$, respectively; their class-conditional frontiers at $\alpha=5\%$ are 25.8\% and 27.3\%.}

\label{tab:frontier}

\setlength{\tabcolsep}{4.5pt}
\renewcommand{\arraystretch}{1.08}

\resizebox{\textwidth}{!}{%
\begin{tabular}{
    l
    c
    c
    r
    c
    r
    r
    r
    r
    r
    r
    r
}
\toprule

\makecell[l]{Model}
&
\makecell{Fam.}
&
\makecell{Adapt.}
&
\makecell{Params.\\(B)}
&
\makecell{NF4}
&
\makecell{$\mathrm{Cov}_{\mathrm{A}}$\\(\%)}
&
\makecell{$\mathrm{Cov}_{\mathrm{R}}$\\(\%)}
&
\makecell{Overall\\coverage (\%)}
&
\makecell{Escalation\\(\%)}
&
\makecell{Amortized\\size (\% BF16)}
&
\makecell{BF16\\VRAM (GB)}
&
\makecell{NF4\\VRAM (GB)}
\\

\midrule

\multicolumn{12}{l}{\textit{\textbf{70--72B models}}}\\[-1mm]
\addlinespace[1mm]

Qwen2.5-72B & Q & G & 72 & \checkmark & \textbf{97.4} & \textbf{100.0} & \textbf{98.9} & 1.1 & \textbf{25.8} & 146.3 & 43.0 \\
HuatuoGPT-o1-72B & Q & D & 72 & \checkmark & \textbf{92.7} & \textbf{100.0} & \textbf{96.9} & 3.1 & \textbf{27.3} & 146.1 & 42.7 \\
Med42-70B & L & D & 70 & \checkmark & 97.7 & 88.2 & 94.2 & 5.8 & \textbf{29.4} & 141.9 & 41.2 \\
OpenBioLLM-70B & L & D & 70 & \checkmark & 98.7 & \emph{0.0} & 92.6 & 7.4 & \textbf{30.6} & 141.7 & 41.0 \\
Llama-3-70B & L & G & 70 & \checkmark & 85.9 & 92.3 & 88.1 & 11.9 & \textbf{33.9} & 141.9 & 41.2 \\
MMed-Llama-3.1-70B & L & D & 70 & \checkmark & 96.5 & \emph{0.0} & 86.4 & 13.6 & \textbf{35.2} & 141.9 & 41.2 \\
Llama-3.3-70B & L & G & 70 & \checkmark & 72.3 & \textbf{100.0} & 81.2 & 18.8 & \textbf{39.1} & 141.9 & 41.2 \\

\midrule

\multicolumn{12}{l}{\textit{\textbf{7--8B models}}}\\[-1mm]
\addlinespace[1mm]

BioMistral-7B & M & D & 7 & \checkmark & 96.3 & \emph{0.0} & 95.7 & 4.3 & \textbf{28.2} & 14.8 & 4.6 \\
Llama-3-8B & L & G & 8 & \checkmark & 89.6 & 94.8 & 91.0 & 9.0 & \textbf{31.7} & 16.7 & 6.4 \\
Mistral-7B & M & G & 7 & \checkmark & 89.4 & 80.6 & 84.9 & 15.1 & 36.3 & 14.9 & 4.7 \\
MMed-Llama-3-8B & L & D & 8 & \checkmark & \textbf{100.0} & 35.1 & 79.8 & 20.2 & 40.2 & 16.7 & 8.7 \\

\bottomrule
\end{tabular}%
}

\vspace{3pt}

\begin{minipage}{0.98\textwidth}
\footnotesize

\textit{Notes.}
\textbf{Family:} architecture family (Q, Qwen; L, Llama; M, Mistral).
\textbf{Adaptation:} General, general-purpose checkpoint; Domain, domain-adapted checkpoint (G, general; D, domain).
\textbf{NF4:} Frontier configuration NF4--4 denotes 4-bit NF4, which is the
lowest-amortized-cost configuration in the evaluated candidate set for all
eleven models at $\alpha=5\%$ under the class-conditional policy.
\textbf{Cov$_{\mathrm{A}}$ / Cov$_{\mathrm{R}}$:} certified coverage within
the predicted ACCEPT and REJECT decision classes.
\textit{0.0} denotes complete escalation of that predicted class, not missing
data or failed inference.
\textbf{Overall coverage:} fraction of held-out decisions served by the
quantized model using the observed test-set class composition.
\textbf{Escalation:} fraction routed to the BF16 reference
(Overall coverage + Escalation = 100\%).
\textbf{Amortized size:} coverage-weighted deployment cost as a percentage
of the 16-bit BF16 reference.
\textbf{VRAM:} measured peak GPU allocation during inference.
For configurations with non-zero escalation, fitting the NF4 compressed model
within 48 GB does not by itself imply a one-GPU end-to-end deployment, because
the BF16 reference must remain reachable for escalated decisions.
The zero-escalation rows are the configurations for which the compressed model
alone can serve the evaluated decision stream.
\end{minipage}
\end{table*}

\begin{figure}[t]
\centering
\includegraphics[width=\textwidth]{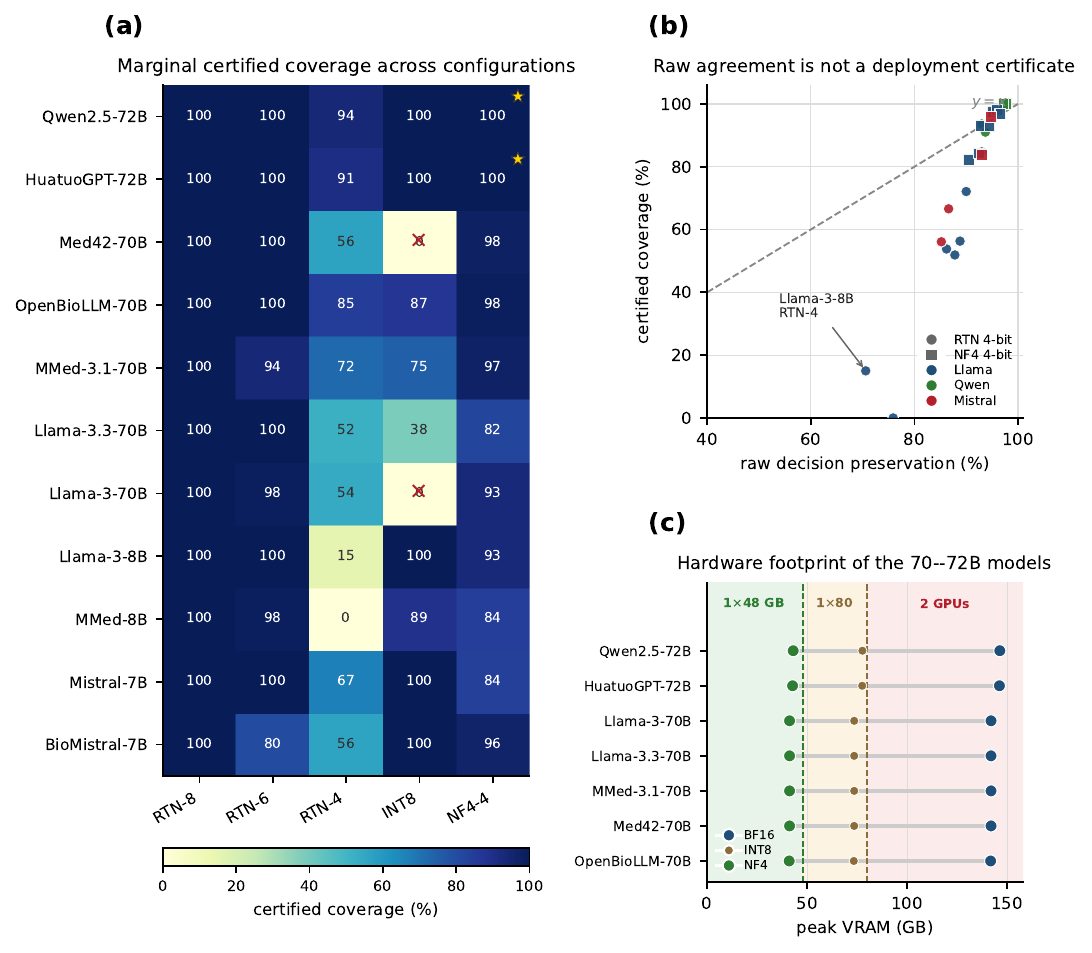}
\caption{\textbf{Decision certification across eleven models, the gap between raw agreement and what
is statistically certifiable, and the resulting hardware footprint.}
\textbf{a}, Marginal certified coverage (\%) for each model under round-to-nearest at $8$, $6$ and $4$ bits,
$8$-bit \texttt{LLM.int8()} (INT8) and NF4 $4$-bit, at $\alpha=5\%$; the class-conditional (Mondrian) policy
used for clinical deployment is shown separately in Fig.~\ref{fig:risk} and Table~\ref{tab:frontier}. Eight-bit round-to-nearest certifies at
least $99.8\%$ for every model, whereas round-to-nearest $4$-bit is strongly model-dependent. Stars
mark configurations that certify at $100\%$ with no escalation; crosses mark the \texttt{LLM.int8()}
configurations that produce degenerate outputs on two BF16-native $70$B checkpoints (an implementation
failure) and are assigned zero deployable coverage.
\textbf{b}, Raw decision preservation versus certified coverage for every model at round-to-nearest
$4$-bit (circles) and NF4 $4$-bit (squares), coloured by family. Points lie below the identity line
because the certificate rejects the low-margin cases that raw agreement would accept; the mean gap
is $14.8$ percentage points (for example Llama-3-8B at round-to-nearest $4$-bit, raw $70.6\%$ but
certified $15.0\%$).
\textbf{c}, Measured peak VRAM for each of the seven $70$--$72$B models at BF16, $8$-bit
\texttt{LLM.int8()} (INT8) and NF4 $4$-bit; the three shaded regions mark how many accelerators the footprint
requires. Every model moves from $\sim$$142$--$146$\,GB at BF16, which spans two $80$\,GB GPUs, to
$\sim$$41$--$43$\,GB under NF4, below the nominal capacity of a single $48$\,GB device (not directly tested on that hardware).}
\label{fig:landscape}
\end{figure}

\begin{figure}[t]
\centering
\includegraphics[width=\textwidth]{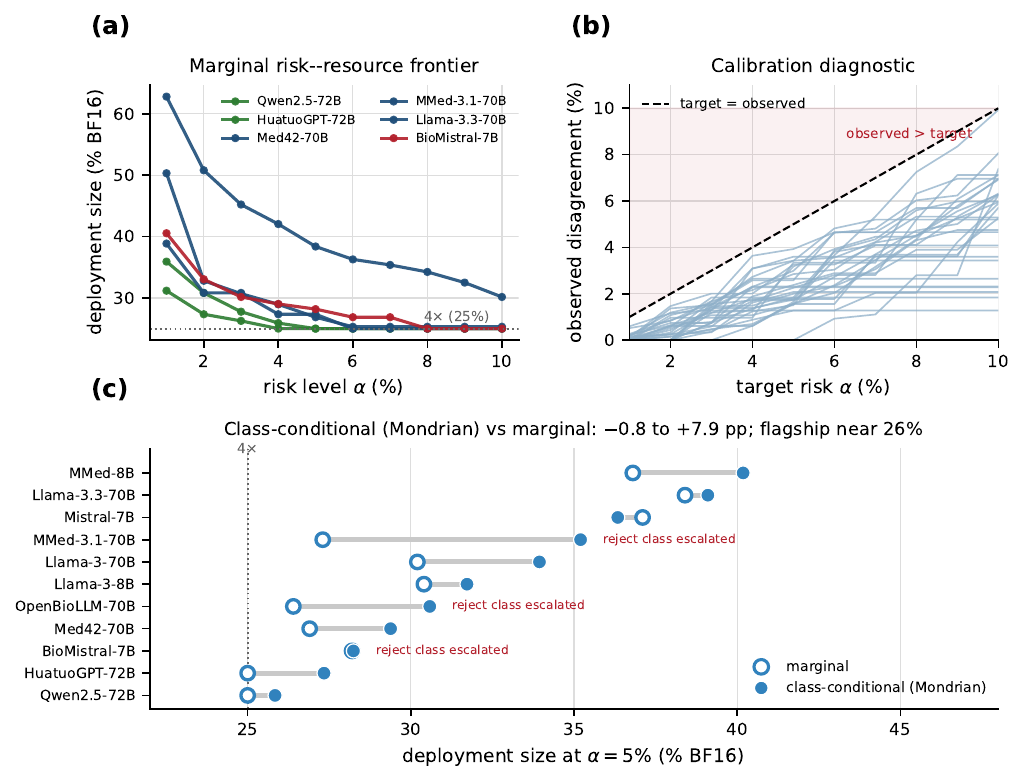}
\caption{\textbf{A risk level determines the deployable footprint, the certificate is calibrated,
and the class-conditional guarantee costs little.}
\textbf{a}, Marginal (lower-envelope) deployment size as a function of the risk level $\alpha$ for six
representative models; every model retains a certified operating point across the range, and the
monotonic, stable curves approach the $4$-bit floor ($25\%$, dotted) as $\alpha$ relaxes.
\textbf{b}, Calibration diagnostic: for every model and quantizer (light lines), the disagreement
observed on the held-out test set at the deployed operating point stays at or below the target
$\alpha$ (dashed identity) across $\alpha$ from $1\%$ to $10\%$; no line enters the shaded region, the
closest approach being Med42-70B under round-to-nearest $4$-bit at $\alpha=10\%$.
\textbf{c}, Deployment size at $\alpha=5\%$ under the marginal certificate versus the class-conditional
(Mondrian) policy. The per-class guarantee changes size by $-0.8$ to $+7.9$ percentage points (median $+2.3$); three
models escalate their entire reject class rather than certify it unsafely (labelled), and the
flagship models remain near $26\%$.}
\label{fig:risk}
\end{figure}

\begin{figure}[t]
\centering
\includegraphics[width=0.92\textwidth]{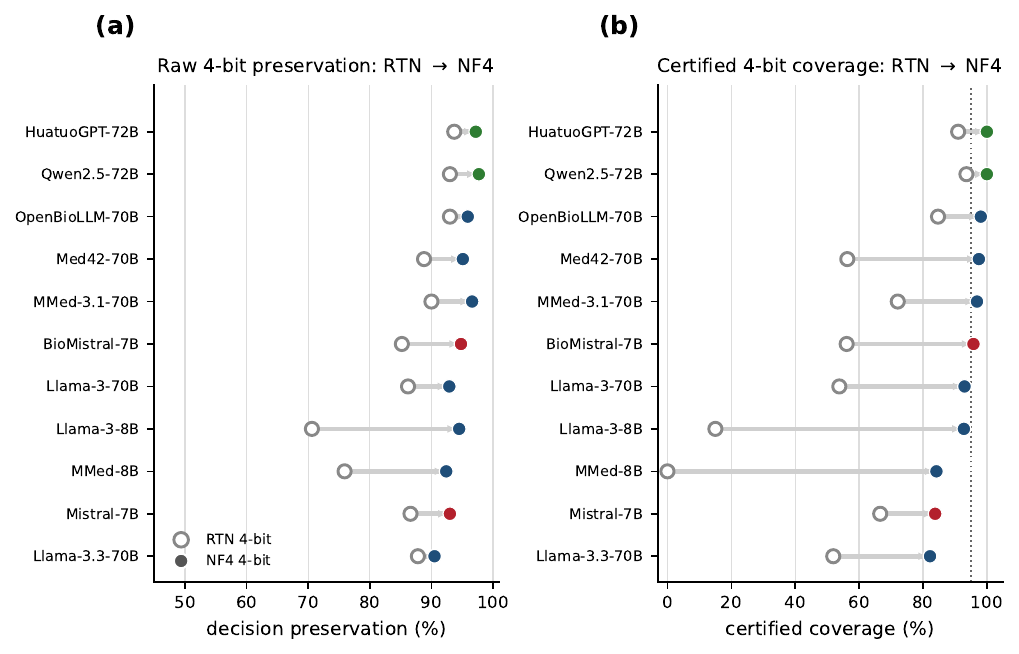}
\caption{\textbf{The quantizer determines whether four-bit compression is certifiable.}
\textbf{a}, Raw $4$-bit decision preservation moving from round-to-nearest (open circles) to NF4
(filled, coloured by family) for each model; NF4 improves preservation for all eleven models.
\textbf{b}, Certified coverage at $\alpha=5\%$ for the same models moving from round-to-nearest
(open circles) to NF4 (filled) at $4$-bit; NF4 lifts certified coverage into the $82.2\%$ to $100\%$
range (dotted line, $95\%$). The domain fine-tunes that round-to-nearest had failed, OpenBioLLM-70B
($98.1\%$), Med42-70B ($97.5\%$), MMed-Llama-3.1-70B ($96.9\%$) and BioMistral-7B ($95.8\%$), all
certify at high coverage under NF4, and MMed-Llama-3-8B rises from $0\%$ to $84.2\%$. Both panels
share the same row order, so a model can be tracked across them.}
\label{fig:quantizer}
\end{figure}

\begin{figure}[t]
\centering
\includegraphics[width=\textwidth]{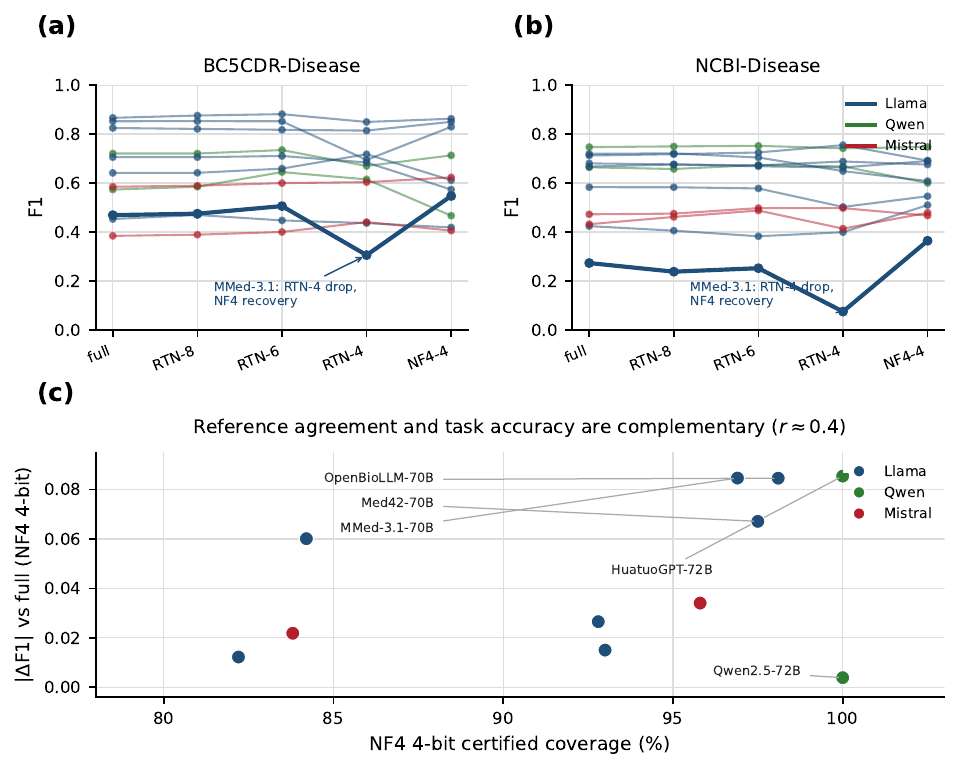}
\caption{\textbf{The frontier-selected configurations preserve external task accuracy, which is complementary
to, not the same as, reference agreement.}
\textbf{a},\textbf{b}, Disease named-entity F1 against human annotations on BC5CDR-Disease
(\textbf{a}) and NCBI-Disease (\textbf{b}) in BF16, round-to-nearest $8/6/4$-bit and the
frontier-selected NF4 $4$-bit configuration. Eight- and six-bit round-to-nearest quantization changes the F1 score negligibly; the frontier-selected NF4 $4$-bit
configurations change it comparably to round-to-nearest $4$-bit quantization (mean absolute change
$0.045$ on BC5CDR and $0.045$ on NCBI), and MMed-Llama-3.1-70B, the worst round-to-nearest $4$-bit
casualty ($-0.163$ and $-0.198$), recovers under the frontier-selected NF4 configuration ($+0.078$ and
$+0.091$).
\textbf{c}, NF4 $4$-bit certified coverage versus the mean absolute change in F1. The per-model
association is weak and the eleven configurations occupy distinct positions: the certificate concerns agreement with the reference on the
verifier task, whereas the benchmark concerns correctness against human labels on a different task,
so the two provide complementary evidence.}
\label{fig:external}
\end{figure}

\bibliography{sn-bibliography}
\clearpage
\end{document}


\maketitle
\vspace{-3em}

\noindent This Supplementary Information provides the formal certificate, extended methods, complete
per-model audit tables, statistical and distributional robustness analyses, and the reproducibility
record for CERTA. It is organized as thirteen sections: study design and task (S1); the formal
finite-sample guarantee with the certification algorithm (S2); the class-conditional derivation (S3);
the complete per-configuration and per-class certificate audit (S4); the two mechanisms by which a
decision class is escalated (S5); statistical robustness to the confidence level, calibration size,
threshold grid and split resampling (S6); distributional robustness to prevalence shift (S7);
deployment behaviour on escalated cases (S8); task-wise certification (S9); external human-gold
validation (S10); the deterministic-bound impossibility result (S11); the reproducibility record of
models, software and commands (S12); and a glossary of the statistical and quantization terms with the
rationale for each (S13). The primary certificate and all main-text results use the
prespecified patient-disjoint split ($n_{\mathrm{cal}}=1{,}250$, $n_{\mathrm{test}}=1{,}250$,
$2{,}484$ patients); the robustness analyses below additionally repeat the procedure over twenty
independent patient-disjoint splits and re-run it with one decision per patient. All certificate and
robustness analyses use the recorded per-input reference decisions, quantized decisions and quantized
margins and require no additional model inference; the external named-entity results are taken from the
locked benchmark outputs. Confidence intervals for empirical clinical-verifier coverage estimates are
patient-clustered bootstrap intervals (patients, not individual decisions, are resampled, respecting the
sixteen patients that contribute two decisions each); confidence intervals for the external
named-entity benchmarks (BC5CDR-Disease, NCBI-Disease), whose evaluation units are public documents
rather than patients, are document-level bootstrap intervals. All bootstraps use $1{,}000$ resamples.
These intervals quantify empirical-estimate uncertainty; the formal decision-risk control is provided by
the Clopper--Pearson certificate described below, not by the bootstrap.

\vspace{0.6em}
\noindent\fbox{\parbox{0.97\textwidth}{\small
\textbf{What CERTA does, and does not, certify.} CERTA controls \emph{compression-induced disagreement
risk}: the rate at which the quantized model's decision differs from the BF16 reference on the inputs
it is permitted to serve. It does \emph{not} control \emph{clinical-correctness risk}: whether the
reference decision is itself correct against a clinical ground truth. The external named-entity
benchmarks estimate a third, separate quantity, \emph{task-performance change} against human labels.
These three risks are kept distinct throughout: the certificate bounds the first, the benchmarks probe
the third, and the second is the object of the future clinician-adjudication study.}}

\vspace{0.6em}
\noindent\textbf{Clinical reading of the decision classes.} An $\YES$ decision retains a candidate
clinical entity or answer; a $\NO$ decision suppresses it. A disagreement with the reference can correspond either
to a false accept or a false reject, whose clinical consequences may differ by workflow. CERTA
therefore does not impose an \emph{a priori} stricter risk on one class: the same target $\alpha$ is
controlled separately for the $\YES$ and $\NO$ classes, and whichever class cannot be certified at that
target is escalated. Empirically it is usually the $\NO$ class that is escalated, because it is
the rarer decision and shows poorer compressed-versus-reference agreement (Supplementary
Tables~\ref{stab:mond} and~\ref{stab:classres}).

\tableofcontents
\vspace{1em}

\section{Study design and task}
Table~\ref{stab:notation} defines the quantities used throughout the paper. (For further explanation of terminology and methodological rationale, see Supplementary Note 13.)

The de-identified clinical narrative
corpus is the publicly available PMC-Patients dataset of patient summaries curated from open-access
PubMed Central case reports; it is sampled into the verifier accept/reject decision set that CERTA
certifies. This set is split patient-disjointly into calibration and held-out test, and the same model
checkpoints are separately evaluated on independent, human-annotated public named-entity benchmarks.

\begin{table}[h]\centering\small
\caption{\textbf{Interpretation of quantities.} Symbols and quantities used in the certificate and
throughout this Supplement.}
\label{stab:notation}
\setlength{\tabcolsep}{8pt}
\renewcommand{\arraystretch}{1.18}
\begin{tabular}{lp{13.3cm}}
\toprule
Symbol & Meaning\\
\midrule
$x$ & a single decision input (a candidate clinical accept/reject decision)\\
$d^{\mathrm{ref}}(x)$ & BF16 reference decision, $\YES$ or $\NO$\\
$\hat d(x)$ & quantized-model decision on the same input\\
$g(x)$ & predicted decision class of the class-conditional certificate, $g(x)=\hat d(x)\in\{\YES,\NO\}$; the two classes are abbreviated $\mathrm{A}$ (accept) and $\mathrm{R}$ (reject) in subscripts\\
$m_k(x)$ & decision margin $|z_{\mathrm{A}}-z_{\mathrm{R}}|$ between the accept and reject answer-token logits at $k$-bit precision\\
$y(x)$ & disagreement indicator $\mathbf{1}[\hat d(x)\neq d^{\mathrm{ref}}(x)]$ (where the indicator function flags disagreements: 1 means disagreement; 0 means agreement).\\
$\alpha$ & target per-decision disagreement risk (risk level)\\
$\delta$ & confidence parameter; the certificate holds with probability $\ge 1-\delta$\\
$G$ & number of candidate thresholds in the calibration grid\\
$\tau,\ \tau_g$ & selected margin threshold (marginal; per predicted class $g$)\\
$n_{\mathrm{cal}},\ n_{\mathrm{cal},g}$ & calibration inputs (total; in predicted class $g$)\\
$S_g(x;\tau_g)$ & class-$g$ service indicator, $\mathbf{1}[g(x)=g,\ m(x)\ge\tau_g]$\\
$\Risk(\tau)$ & marginal disagreement risk on the served region, $\Prob(\hat d\neq d^{\mathrm{ref}}\mid m\ge\tau)$\\
$\Risk_g(\tau_g)$ & class-$g$ disagreement risk, $\Prob(\hat d\neq d^{\mathrm{ref}}\mid g(x)=g,\ m\ge\tau_g)$\\
$U_{\mathrm{CP}}(\tau)$ & one-sided Clopper--Pearson upper confidence bound on the risk at threshold $\tau$\\
$\alpha_{\min,g}$ & zero-disagreement resolution floor $1-(\delta/2G)^{1/n_{\mathrm{cal},g}}$\\
$k$ & nominal bit-width of the compressed configuration (e.g. $4$ for NF4)\\
$C$ & coverage: fraction of decisions served by the quantized model\\
$E$ & escalation: fraction routed to the BF16 reference ($E=1-C$)\\
$B_{\mathrm{amortised}}$ & amortized size: coverage-weighted nominal-bit cost as a percentage of BF16, $100[1-(1-k/16)C]$; a BF16-equivalent nominal-bit metric, not physical VRAM (\% BF16)\\
\bottomrule
\end{tabular}
\end{table}

The certified decision is the accept/reject step of an in-house clinical information-extraction
and verification pipeline that we have not yet published; only this accept/reject step is used or
evaluated here. Once the verifier decisions and quantized margins have been recorded, the certification
procedure itself does not depend on any pipeline-specific model internals; its distributional validity
depends only on how the inputs were sampled and on the exchangeability of calibration and test
decisions. In this pipeline, a generator model proposes candidate clinical entities and answers extracted from
de-identified patient narratives, and a verifier model adjudicates each candidate. The evaluated
decision space is binary, $\YES$ and $\NO$, and the model is constrained to a
single-token accept/reject answer rendered with each checkpoint's native chat template; for checkpoints whose
template does not accept a system role, the system instruction is merged into the user turn. Two
candidate types were evaluated in equal numbers: \emph{named-entity verification} (``Is a candidate
entity, with a proposed category, a valid and specific clinical entity of that category?'') and
\emph{answer verification} (``Is a proposed answer supported by the supplied context and question?'').
Reject cases were constructed by substituting a mismatched answer or a different clinical category, so
that the evaluation contains controlled reject cases (constructed from real clinical contexts) rather
than an accept-only distribution.

\paragraph{Decision distribution.} Because class imbalance in the verifier's own decisions is what
motivates the class-conditional policy, Supplementary Table~\ref{stab:decdist} reports, for every
model, the prevalence of $\YES$/$\NO$ decisions in the BF16 reference and in the NF4 quantized model,
overall and separately for the two task types. Several clinically adapted models are strongly
acceptance-skewed (for example, the reject class of MMed-Llama-3.1-70B and OpenBioLLM-70B is a small
minority of their decisions), which is precisely the regime in which a single marginal threshold can
leave the rare reject decision poorly controlled.

\begin{table}[h]\centering\small
\caption{\textbf{Decision distribution for various models.} Prevalence (\%) of accept ($\mathrm{A}$) and reject ($\mathrm{R}$)
decisions for the BF16 reference and the NF4 4-bit model overall and split by task
(named-entity / answer verification), on the full evaluation set.}
\label{stab:decdist}
\setlength{\tabcolsep}{5pt}
\begin{tabular}{lrrrrcc}
\toprule
Model & \makecell{Ref\\A} & \makecell{Ref\\R} & \makecell{NF4\\A} & \makecell{NF4\\R}
& \makecell{Ref A\\NER/QA} & \makecell{NF4 A\\NER/QA}\\
\midrule
\tableSdecdist
\bottomrule
\end{tabular}
\end{table}

\section{Formal CERTA guarantee}
Algorithm~\ref{alg:certa} states the calibration procedure, an instance of \emph{Learn-then-Test}: a
distribution-free calibration method that fixes a finite grid of candidate thresholds and then selects
one on held-out data so that a user-specified risk is controlled with finite-sample, high-probability
confidence (main-text Methods). To keep the candidate grid independent of
the data used to certify it, CERTA splits the calibration decisions into a \emph{threshold-selection}
subset, used only to construct the $G$-point candidate grid, and a disjoint \emph{certification} subset,
used only to evaluate the Clopper--Pearson bounds; coverage is then measured on the held-out test set.
Because the grid is fixed before the certification labels are read, the finite-sample Learn-then-Test
guarantee is exact: at each fixed candidate threshold, the served certification decisions form a binomial
disagreement sample, and a one-sided Clopper--Pearson upper bound at the Bonferroni-corrected level
$1-\delta/G$ controls its risk (Algorithm~\ref{alg:certa}, Lines~\ref{ln:count}--\ref{ln:cp}), and the smallest threshold whose bound clears the target is deployed (Line~\ref{ln:select}; a
class with no admissible threshold is escalated in full). The class-conditional variant runs the same
procedure independently within each predicted decision class at $1-\delta/(2G)$.

The certificate reported
in the main text uses the full $1{,}250$-decision calibration set for both the grid and the bounds, a
higher-power operational estimate. The exact certificate instead handles every source of data-dependent
selection: it builds the grid on the $250$-decision selection subset (Lines~\ref{ln:grid}--\ref{ln:alloc}) and evaluates the bounds on the
$1{,}000$-decision certification subset reduced to one decision per patient ($996$ decisions). The selection,
certification and $1{,}250$-decision test subsets are mutually patient-disjoint, so no patient appears in more
than one (the $2{,}500$ decisions come from $2{,}484$ patients, each confined to a single subset), the grid is
independent of the certification labels, and, conditional on the number of served certification inputs
$n_\tau$, the disagreement count $x_\tau$ is binomial under the patient-level iid assumption; and it applies an outer $\delta/H$ Bonferroni across the $H$ deployable
configurations, so the guarantee holds for the configuration the frontier selects
(Proposition~\ref{prop:cert} and Remark~\ref{rem:config}). The exactness check comparing this frontier
with the reported one is given in the statistical-robustness section (Supplementary
Table~\ref{stab:exact}).

\begin{algorithm}[t]
\caption{\textbf{CERTA certify-or-escalate calibration} (one decision class; run once per class for the
class-conditional variant).}
\label{alg:certa}
\small
\begin{algorithmic}[1]
\Require threshold-selection set $\mathcal{S}_g$ and certification set $\mathcal{C}_g$ (disjoint calibration
subsets, each restricted to inputs whose quantized prediction is class $g$), each a set of pairs $(m_i,y_i)$ with margin $m_i$ and disagreement flag
$y_i=\mathbf{1}[\hat d_i\neq d^{\mathrm{ref}}_i]$; target risk $\alpha$; confidence allocation $\delta'$; grid size $G$. \emph{Marginal variant:} use the unrestricted sets $\mathcal{S}$, $\mathcal{C}$. \emph{Allocation:} set $\delta'=\delta/G$ (marginal, one configuration) or $\delta/(2G)$ (class-conditional, one configuration); for the exact frontier selected among $H$ deployable configurations, $\delta'=\delta/(GH)$ or $\delta/(2GH)$ respectively.
\Ensure threshold $\tau^\star$ and a serve-or-escalate rule
\If{$\mathcal{S}_g=\emptyset$} \State $\tau^\star_g\gets\infty$ \textbf{and return} \Comment{no class-$g$ selection data: escalate class $g$ in full} \EndIf
\State $\mathcal{T}\gets G$-point grid of empirical margin quantiles of $\mathcal{S}_g$ \Comment{fixed before $\mathcal{C}_g$ is read}\label{ln:grid}
\State use the supplied per-threshold allocation $\delta'$ \Comment{$\delta/G$ or $\delta/(2G)$ for one configuration; $\delta/(GH)$ or $\delta/(2GH)$ for the exact configuration-selected frontier}\label{ln:alloc}
\For{each $\tau\in\mathcal{T}$}\label{ln:for}
  \State $n_\tau\gets|\{i\in\mathcal{C}_g:m_i\ge\tau\}|$,\quad $x_\tau\gets|\{i\in\mathcal{C}_g:m_i\ge\tau,\ y_i=1\}|$ \Comment{marginal variant: use unrestricted $\mathcal{C}$}\label{ln:count}
  \State $U_{\mathrm{CP}}(\tau)\gets
  \begin{cases}
  1, & n_\tau=0 \text{ or } x_\tau=n_\tau,\\
  \mathrm{Beta}^{-1}_{\,1-\delta'}(x_\tau+1,\ n_\tau-x_\tau), & \text{otherwise}
  \end{cases}$ \Comment{Clopper--Pearson upper bound}\label{ln:cp}
\EndFor
\State $\tau^\star_g\gets \min\{\tau\in\mathcal{T}:U_{\mathrm{CP}}(\tau)\le\alpha\}$, or $\tau^\star_g\gets\infty$ if no such $\tau$ \Comment{$\infty$ escalates class $g$ in full}\label{ln:select}
\State \textbf{Deployment:} serve input $x$ with the quantized model if $m_k(x)\ge\tau^\star_{\widehat g(x)}$ (class-conditional; $m_k(x)\ge\tau^\star$ for the marginal variant, which uses a single threshold), otherwise escalate $x$ to the BF16 reference\label{ln:deploy}
\end{algorithmic}
\end{algorithm}

For the class-conditional variant, the allocation on Line~\ref{ln:alloc} becomes $\delta'=\delta/(2G)$; the
exact $H$-configuration frontier uses $\delta/(GH)$ or $\delta/(2GH)$ (Remark~\ref{rem:config}). Here
$\mathrm{Beta}^{-1}_{\,q}(a,b)$ denotes the $q$-quantile of the $\mathrm{Beta}(a,b)$ distribution,
equivalently the inverse of the regularized incomplete beta function, which is the standard closed form
of the one-sided Clopper--Pearson upper confidence bound. For the boundary cases used in the algorithm,
$U_{\mathrm{CP}}(\tau)$ is defined as $1$ when $n_\tau=0$ or $x_\tau=n_\tau$; otherwise it is the
Beta-quantile expression below.

\begin{proposition}[Finite-sample decision-risk control]\label{prop:cert}
Fix $\alpha,\delta\in(0,1)$ and let the candidate grid $\mathcal{T}$ ($|\mathcal{T}|\le G$) be fixed
independently of the certification set, and suppose the certification set contains one decision per
patient, with these certification patients and future deployment patients independent and identically
distributed (so the served certification units are independent). For each $\tau\in\mathcal{T}$ let $x_\tau$ and $n_\tau$ be the number of disagreements and
the number of served inputs in the certification set, and let
$\Risk(\tau)=\Prob(\hat d\neq d^{\mathrm{ref}}\mid m\ge\tau)$ denote the served-region disagreement risk.
The one-sided Clopper--Pearson upper bound
\begin{equation*}
U_{\mathrm{CP}}(\tau)=
\begin{cases}
1, & n_\tau=0 \ \text{or}\ x_\tau=n_\tau,\\[2pt]
\mathrm{Beta}^{-1}_{\,1-\delta/G}(x_\tau+1,\,n_\tau-x_\tau), & \text{otherwise,}
\end{cases}
\end{equation*}
is valid for $\Risk(\tau)$. Selecting the smallest $\tau^\star$ with $U_{\mathrm{CP}}(\tau^\star)\le\alpha$ gives,
with probability at least $1-\delta$ over the draw of the certification sample, a disagreement risk
$\Risk(\tau^\star)\le\alpha$ whenever a finite threshold is selected; if no admissible threshold exists the
procedure escalates the class in full ($\tau^\star=\infty$) and serves no compressed decisions.
Running the procedure separately within each predicted class with allocation $\delta/(2G)$ yields
\begin{equation*}
\Prob\!\left(\bigcap_{g\in\{\mathrm{A},\mathrm{R}\}}\left\{\tau_g^\star<\infty \ \Longrightarrow\ \Risk_g(\tau_g^\star)\le\alpha\right\}\right)\ge 1-\delta ,
\end{equation*}
that is, every class served by the compressed model satisfies the target risk, while a fully escalated
class ($\tau_g^\star=\infty$) serves no compressed decisions and contributes no compression-induced
disagreement; here $\Risk_g(\tau_g)$ is the class-conditional disagreement risk within the predicted
accept ($\mathrm{A}$) or reject ($\mathrm{R}$) class, defined in Table~\ref{stab:notation}. The guarantee
is stated for the sampling unit of one decision drawn per independent patient according to the deployment
sampling protocol; it does not automatically extend to an unrestricted collection of multiple correlated
decisions from the same patient, which would require a cluster-level sampling or risk-control argument.
\end{proposition}

\noindent\emph{Plain-language reading.} Try the candidate margin thresholds in turn. At each one, count how
often the quantized model disagrees with the BF16 reference on the cases it would serve, and compute a
one-sided confidence bound $U_{\mathrm{CP}}$ on that disagreement rate. Deploy the smallest threshold whose
bound is already at or below the target $\alpha$; if none qualifies, escalate the class in full. Because the
grid is fixed before the certification labels are read and the bound is widened to cover all $G$ thresholds
tried, the served decisions provably disagree with the reference at most a fraction $\alpha$ of the time,
with confidence $1-\delta$.

\begin{proof}[Proof sketch]
Fix $\tau\in\mathcal{T}$. Since $\mathcal{T}$ is built from the selection set, which is disjoint from the
certification set, $\tau$ is independent of the certification labels; conditioning on the number of
served certification inputs $n_\tau$, the served-set disagreement count $x_\tau$ is
$\mathrm{Binomial}(n_\tau,\Risk(\tau))$ under the one-decision-per-patient iid assumption, so on the
non-boundary thresholds ($n_\tau>0$ and $x_\tau<n_\tau$) the one-sided Clopper--Pearson quantity
$U_{\mathrm{CP}}(\tau)=\mathrm{Beta}^{-1}_{\,1-\delta/G}(x_\tau+1,\,n_\tau-x_\tau)$ is a valid $1-\delta/G$ upper confidence
bound: $\Prob(\Risk(\tau)>U_{\mathrm{CP}}(\tau))\le\delta/G$. For the boundary thresholds ($n_\tau=0$ or
$x_\tau=n_\tau$) the bound is defined as $U_{\mathrm{CP}}(\tau)=1$; since $\alpha\in(0,1)$ such a threshold is
never admissible, so it is never selected and the beta-quantile argument is only needed on the
non-boundary cases. A Bonferroni union bound over the $|\mathcal{T}|\le G$
fixed candidates makes the probability that any candidate's bound understates its risk at most $\delta$;
on the complementary event, if an admissible threshold exists, the selected $\tau^\star$, which satisfies $U_{\mathrm{CP}}(\tau^\star)\le\alpha$ by
construction, obeys $\Risk(\tau^\star)\le U_{\mathrm{CP}}(\tau^\star)\le\alpha$, giving
$\Prob(\Risk(\tau^\star)\le\alpha)\ge 1-\delta$ for the (nonempty) served region. When no admissible
threshold exists the class is escalated in full ($\tau^\star=\infty$): no compressed decisions are served,
the conditional risk is not defined, and the class contributes no compression-induced deployment
disagreement, so the all-escalated outcome is consistent with the policy. For the class-conditional variant, the same argument
applied within each predicted class at allocation $\delta/(2G)$ gives each class a $1-\delta/2$ guarantee,
and a union bound over the two classes yields the joint statement.
\end{proof}

\begin{remark}[Selection across configurations]\label{rem:config}
When the deployed configuration is chosen by minimum amortized size over $H$ candidate deployable
configurations, the choice is itself data-dependent. Replacing the per-threshold allocation $\delta/G$ by
$\delta/(GH)$ (and $\delta/(2G)$ by $\delta/(2GH)$ for the class-conditional variant) and taking a union
bound over the $H\,|\mathcal{T}|$ fixed (configuration, threshold) pairs makes the guarantee hold
simultaneously for the selected configuration. The exact certificate uses this correction with $H=2$
deployable configurations (NF4 and $8$-bit \texttt{LLM.int8()}); round-to-nearest retains BF16 storage and is
excluded from deployment selection. Consistent with the main text, the configuration is selected using
calibration-side coverage, so the selection is independent of the test set, which is used only to estimate
the selected configuration's coverage and amortized cost (an out-of-sample estimate). The higher-power
operational frontier reported in the main text omits this outer factor; Supplementary Table~\ref{stab:exact} shows that reinstating it, together with the
independent grid and one-decision-per-patient certification, changes the frontier by a median of $1.0$ and
at most $8.2$ percentage points.
\end{remark}

\noindent\emph{Plain-language reading.} Choosing which compressed configuration to deploy on the basis of
its cost is itself a decision made from the data, so we must pay for it statistically: the confidence budget
is split across the $H$ candidate configurations as well as the $G$ thresholds, which is why $\delta/G$
becomes $\delta/(GH)$. With only $H=2$ deployable configurations, this costs almost nothing, and the
main-text frontier, which skips this extra split for statistical power, lands within a median of $1.0$
percentage point of the fully corrected one.

\section{Class-conditional certification}
The marginal certificate selects a single threshold $\tau$ by Learn-then-Test over a grid of
$G=50$ candidate thresholds, using a Bonferroni allocation $\delta/G$ so that the finite-sample
disagreement bound $R_k(\tau)\le\alpha$ holds with confidence at least $1-\delta$. When two
class-conditional certificates are calibrated jointly (one per predicted decision class), controlling
each class independently at confidence $1-\delta$ would guarantee only $1-2\delta$ jointly. To retain
the stated $1-\delta$ over both classes, each class receives a Bonferroni allocation of $\delta/(2G)$
across its own threshold grid. For a single, prespecified configuration whose candidate grid is fixed
independently of the certification sample, a union bound over the two class-conditional events then yields
the joint class-conditional guarantee of Proposition~\ref{prop:cert}: with probability at least $1-\delta$,
every class served by the compressed model satisfies the target risk, while a fully escalated class
($\tau_g^\star=\infty$) serves no compressed decisions. The higher-power operational class-conditional
analyses in the main text and this Supplement use this $\delta/(2G)$ allocation; the exact split-sample
frontier additionally accounts for selection among the $H=2$ deployable configurations, using
$\delta/(2GH)$ (Remark~\ref{rem:config}, Supplementary Table~\ref{stab:exact}). Unless explicitly stated
otherwise, the operational frontier and its robustness analyses are reported as empirical operating-point
estimates, and the formal finite-sample guarantee applies to the exact split-sample procedure. The
$\delta/(2G)$ correction is inexpensive, and it is a
\emph{distinct} comparison from the marginal-versus-class-conditional cost reported in the main text:
relative to the naive class-wise $\delta/G$ allocation, the joint $\delta/(2G)$ correction increases
each model's class-conditional frontier by at most $2.3$ percentage points of BF16 size, whereas the
main-text figure of $-0.8$ to $+7.9$ points (median $+2.3$) is the cost of the class-conditional
policy relative to the \emph{marginal} certificate. The two flagship models are unaffected by this
correction: at $\alpha=5\%$ their class-conditional deployment sizes are $25.8\%$ (Qwen2.5-72B) and
$27.3\%$ (HuatuoGPT-o1-72B) of the BF16 size under the joint $\delta/(2G)$ allocation, the same values
as under the naive per-class $\delta/G$ allocation (Supplementary Table~\ref{stab:mondalpha}).

\paragraph{Sensitivity of the class-conditional frontier.} The class-conditional frontier inherits the
modest sensitivity of the marginal certificate to the confidence level (Supplementary
Table~\ref{stab:conf}, Mondrian columns): the flagship operating points move by less than one
percentage point of BF16 size (Qwen2.5-72B $25.8\%$ to $26.2\%$, HuatuoGPT-o1-72B $27.3\%$ to
$28.2\%$) as confidence tightens from $90\%$ to $99\%$. Supplementary Table~\ref{stab:mondalpha}
reports the class-conditional frontier and its escalation fraction at $\alpha\in\{1,5,10\}\%$. Relaxing
the risk level moves each model from a high-escalation regime (at $\alpha=1\%$ Qwen2.5-72B certifies
at $59.2\%$ of BF16 size and escalates $45.6\%$ of decisions) down toward the $4$-bit floor (at
$\alpha=10\%$ Qwen2.5-72B and HuatuoGPT-o1-72B reach $25.0\%$ with no escalation), with the deployed
$\alpha=5\%$ operating band ($25.8\%$ to $40.2\%$ across the eleven models) in between.

\section{Complete certificate audit}
Supplementary Table~\ref{stab:perconfig} gives the marginal certified coverage at $\alpha=5\%$ for every
model and configuration (round-to-nearest at $8$, $6$ and $4$ bits, $8$-bit \texttt{LLM.int8()} and NF4
$4$-bit). Supplementary Table~\ref{stab:audit} is the per-operating-point decision-risk audit for the
marginal NF4 points in the higher-power operational analysis; every operating-point Clopper--Pearson
upper bound is at or below the $5\%$ target. Because this analysis builds the threshold grid and
evaluates the bounds on the same full calibration sample, these per-operating-point bounds are
diagnostic rather than an independent finite-sample guarantee; the formal guarantee for the
data-selected deployment frontier is the split-sample certificate of Supplementary
Table~\ref{stab:exact}, which uses an independent threshold-selection subset, one decision per patient
in the certification subset, and a Bonferroni correction across the deployable configurations.
Supplementary Table~\ref{stab:mond} is the
per-class audit at every class-conditional frontier point, reporting each class certificate ($n_g$,
$x_g$, $\tau_g$, coverage and the Clopper--Pearson bound) so that it can be reconstructed independently.
The columns of each table are defined in its caption.

\begin{table}[h]\centering\small
\caption{\textbf{Per-configuration marginal certificate.} Each cell is certified coverage (\%) with a
patient-clustered $95\%$ bootstrap confidence interval, at $\alpha=5\%$, $\delta=0.10$. The point
estimate is the certified test coverage and the brackets are patient-clustered $95\%$ bootstrap
intervals for that coverage; the operating-point Clopper--Pearson upper bound (Table~\ref{stab:audit})
is the operational risk diagnostic rather than the bootstrap interval, while the formal finite-sample
guarantee for the selected frontier is the split-sample certificate (Table~\ref{stab:exact}). INT8 inference failed on two BF16-native $70$B checkpoints
(Llama-3-70B and Med42-70B); these configurations are treated as infeasible and assigned zero compressed
coverage, which is distinct from a statistical certification failure on a successful quantized run.}
\label{stab:perconfig}
\setlength{\tabcolsep}{4pt}
\resizebox{\textwidth}{!}{%
\begin{tabular}{lccccc}
\toprule
Model & RTN-8 & RTN-6 & RTN-4 & INT8 & NF4 \\
\midrule
\tableSperconfig
\bottomrule
\end{tabular}}
\end{table}

\begin{table}[h]\centering\small
\caption{\textbf{Decision-risk audit for the marginal NF4 operating points ($\alpha=5\%$; higher-power
operational analysis).}
The deployed clinical frontier is class-conditional; its per-class audit is Table~\ref{stab:mond}.
$n_{\mathrm{cal}}$: certified calibration inputs ($m_k\ge\tau$); disag.: those disagreeing with the
BF16 reference; $\tau$: selected margin threshold; $U_{\mathrm{CP}}$: one-sided Clopper--Pearson upper bound on
the certified-set disagreement risk (all $\le5\%$); coverage: certified test coverage. The
Clopper--Pearson upper bound here is the operational risk diagnostic: the threshold grid and the bound
are computed on the same full calibration sample, so it is not an independent finite-sample guarantee,
which is instead established by the split-sample certificate (Table~\ref{stab:exact}). The coverage and
its $95\%$ bootstrap CI are an empirical deployment estimate.}
\label{stab:audit}
\setlength{\tabcolsep}{7pt}
\begin{tabular}{lrrrrl}
\toprule
Model & $n_{\mathrm{cal}}$ & disag. & $\tau$ & \makecell{$U_{\mathrm{CP}}$\\(\%)} & coverage (\%) [95\% CI] \\
\midrule
\tableSaudit
\bottomrule
\end{tabular}
\end{table}

\begin{table}[h]\centering\small
\caption{\textbf{Class-conditional (Mondrian) per-class decision-risk audit at the frontier
($\alpha=5\%$, $\delta/(2G)$; higher-power operational analysis).} For each decision class: $n$, certified calibration inputs;
$x$, disagreements among them; $\tau$, selected margin threshold ($\infty$ = class escalated in full);
Cov, certified test coverage; $U_{\mathrm{CP}}$, one-sided Clopper--Pearson upper bound on the class risk (all
$\le5\%$). \emph{esc}: reject class escalated. Each class certificate is reconstructible from
$n$, $x$ and $\tau$. As in Table~\ref{stab:audit}, the threshold grid and the Clopper--Pearson bounds
use the same full calibration sample, so these per-class bounds are operational diagnostics rather than
independent finite-sample certificates; the exact split-sample per-class certificate (independent
selection subset, one decision per patient, $\delta/(2GH)$ allocation) is Table~\ref{stab:exactaudit}.}
\label{stab:mond}
\setlength{\tabcolsep}{4pt}
\resizebox{\textwidth}{!}{%
\begin{tabular}{lrrrrr rrrrr}
\toprule
& \multicolumn{5}{c}{Accept ($\mathrm{A}$)} & \multicolumn{5}{c}{Reject ($\mathrm{R}$)}\\
\cmidrule(lr){2-6}\cmidrule(lr){7-11}
Model & $n_{\mathrm{A}}$ & $x_{\mathrm{A}}$ & $\tau_{\mathrm{A}}$ & Cov (\%) & $U_{\mathrm{CP}}$ (\%)
      & $n_{\mathrm{R}}$ & $x_{\mathrm{R}}$ & $\tau_{\mathrm{R}}$ & Cov (\%) & $U_{\mathrm{CP}}$ (\%)\\
\midrule
\tableSmond
\bottomrule
\end{tabular}}
\end{table}

\section{Why classes are escalated}
Whether a decision class can be certified at the target risk is governed by two distinct conditions,
and a class is escalated if it fails \emph{either}. The first is a resolution condition. With $x=0$
disagreements among $n_{\mathrm{cal},g}$ class-$g$ calibration inputs, the tightest certifiable class
risk is the zero-disagreement floor $\alpha_{\min,g}=1-(\delta/2G)^{1/n_{\mathrm{cal},g}}$; when
$\alpha_{\min,g}>\alpha$ the class has too few calibration inputs to reach the target under any
threshold and is escalated in full. Several clinically adapted models rarely return $\NO$, so their
reject class is a small minority of the calibration set (Supplementary Table~\ref{stab:classres}):
BioMistral-7B has only $4$ reject-class calibration inputs, giving a floor of $82.2\%$, and
OpenBioLLM-70B has $91$, giving a floor of $7.3\%$; both exceed the $5\%$ target, so their reject class
is escalated as \emph{resolution-limited}. The second is a risk condition. Even when the floor lies
below $\alpha$, the smallest Clopper--Pearson upper bound attainable over the threshold grid can still
exceed $\alpha$, because too many of the reject-class calibration decisions disagree with the reference
at every admissible threshold; the class is then escalated as \emph{risk-limited}.
MMed-Llama-3.1-70B is the illustrative case: its reject-class floor is $4.7\%$, below the target, yet
its best attainable Clopper--Pearson bound is $8.4\%$, so it is escalated despite having adequate
resolution. The zero-disagreement floor is therefore a \emph{necessary} resolution condition for
certification, not a \emph{sufficient} one. Supplementary Figure~\ref{sfig:classres} separates the two
mechanisms: every escalated model's reject class falls in either the resolution-limited region (floor
above $\alpha$) or the risk-limited region (floor below $\alpha$ but best attainable bound above it),
while the certified models occupy the low-floor, low-bound corner. This is the statistical reason for
the escalation behaviour in main-text Table~1, not an arbitrary choice.

\begin{table}[h]\centering\small
\caption{\textbf{Class-specific calibration resolution and escalation mechanism for the higher-power operational NF4 frontier.}
$n_{\mathrm{A}}$, $n_{\mathrm{R}}$: calibration inputs in each predicted class; $\alpha_{\min}$: zero-disagreement
resolution floor $1-(\delta/2G)^{1/n}$; min.\ $U_{\mathrm{CP},\mathrm{R}}$: smallest Clopper--Pearson upper bound on the
reject-class risk attainable over the threshold grid; mechanism: whether the reject class is
\emph{certified}, \emph{resolution-limited} (floor $>\alpha$), or \emph{risk-limited} (floor
$\le\alpha$ but best attainable bound $>\alpha$). Target $\alpha=5\%$. The floors use the operational
$\delta/(2G)$ allocation on the full calibration sample, without the configuration-selection factor $H$;
this is an operational diagnostic, distinct from the exact split-sample certificate
(Table~\ref{stab:exactaudit}).}
\label{stab:classres}
\setlength{\tabcolsep}{6pt}
\begin{tabular}{lrrrrrl}
\toprule
Model & $n_{\mathrm{A}}$ & $n_{\mathrm{R}}$ & \makecell{$\alpha_{\min,\mathrm{A}}$\\(\%)} & \makecell{$\alpha_{\min,\mathrm{R}}$\\(\%)}
& \makecell{min.\ $U_{\mathrm{CP},\mathrm{R}}$\\(\%)} & mechanism\\
\midrule
\tableSclassres
\bottomrule
\end{tabular}
\end{table}

\section{Statistical robustness}
\paragraph{Zero-disagreement floor.} With zero observed disagreements among $n_{\mathrm{cal}}$
calibration inputs, the tightest certifiable risk at confidence $1-\delta/G$ is
$1-(\delta/G)^{1/n_{\mathrm{cal}}}$, which at $\delta=0.10$, $G=50$, $n_{\mathrm{cal}}=1{,}250$ is
$\approx 0.5\%$; a $1\%$ target is therefore statistically reachable at this calibration size, and
targets below $\approx 0.5\%$ are not.

\paragraph{Confidence-level sensitivity.} Supplementary Table~\ref{stab:conf} and Supplementary
Figure~\ref{sfig:robust}a report the marginal frontier at $\alpha=5\%$ for
$\delta\in\{0.10,0.05,0.01\}$ (confidence $90\%$, $95\%$, $99\%$), alongside the class-conditional
frontier at the same confidence levels. Across the evaluated confidence levels, the marginal frontier changed only
modestly: the range moves from $25.0$--$38.4\%$ at $\delta=0.10$ to $25.0$--$39.5\%$ at $\delta=0.01$,
and the median stays at $28.2\%$. The headline $25$--$40\%$ deployment envelope therefore does not depend on the choice of a
$90\%$ rather than a $95\%$ or $99\%$ confidence level.

\paragraph{Calibration-size sensitivity.} Supplementary Table~\ref{stab:calsize} and Supplementary
Figure~\ref{sfig:robust}b report the frontier as the calibration set is reduced from $1{,}250$ to
$250$ patient-disjoint decisions. The certificate degrades gracefully, and the aggregate maximum
frontier size increases monotonically as the calibration set is reduced: the least
compressible model moves from $38.4\%$ at $n_{\mathrm{cal}}=1{,}250$ to $49.5\%$ at
$n_{\mathrm{cal}}=250$, while the median across models stays near $28\%$ until the smallest set and the
$4$-bit floor of $25\%$ is retained by the strongest models throughout. More calibration data tightens the frontier, confirming that the $1\%$-resolution floor
is a statistical property of the calibration size rather than a model failure.

\paragraph{Threshold-grid sensitivity.} Learn-then-Test discretizes the margin axis into $G$ candidate
thresholds. Supplementary Table~\ref{stab:grid} and Supplementary Figure~\ref{sfig:gridsplit}a report
the marginal frontier as $G$ is swept over $\{25,50,100,200\}$. The frontier is essentially invariant:
the least-compressible model stays within $38.4$--$39.1\%$ of BF16 size, the median across models moves
only between $28.2\%$ and $29.4\%$, and the strongest models remain pinned at the $25\%$ $4$-bit floor
at every grid size. A finer grid tightens the per-candidate Bonferroni allocation ($\delta/G$) but also
supplies more admissible thresholds; in the evaluated range these changes in threshold availability and
Bonferroni allocation produced little net change in the frontier, so $G=50$ is not a sensitive design
choice.

\paragraph{Repeated patient-disjoint splits.} A single calibration/test split could understate
variability. Supplementary Table~\ref{stab:split} and Supplementary Figure~\ref{sfig:gridsplit}b repeat
the entire certification over $20$ independent patient-disjoint splits, each re-shuffling the $2{,}484$
patient groups before the $50/50$ cut. The marginal frontier is stable across resamples: Qwen2.5-72B
sits exactly at the $25\%$ floor on all $20$ splits and HuatuoGPT-o1-72B within $0.8$ points of it,
while even the most variable model has an interquartile width of at most $2.5$ percentage points of
BF16 size. These results indicate that the observed $25$--$40\%$ deployment envelope is not driven by
the particular patient-disjoint split used for the primary analysis.

\paragraph{One decision per patient.} Sixteen of the $2{,}484$ patients contribute a second decision, so
the $2{,}500$-decision analysis carries a small amount of within-patient dependence. Re-running the
entire certification with one decision per patient (a single decision retained for each of the $2{,}484$
patients, split patient-disjointly) reproduces every frontier: the marginal frontier changes by at most
$1.8$ percentage points of BF16 size (MMed-Llama-3-8B) and the class-conditional frontier by at most
$1.1$ points (Llama-3-70B), with the flagship operating points essentially unchanged (Qwen2.5-72B
$25.8\%$, HuatuoGPT-o1-72B $27.3\%$ to $27.6\%$). The primary analysis uses all $2{,}500$ decisions; the
one-decision-per-patient run is the formally patient-level exchangeable version of the procedure, on which
the exact formal guarantee relies. It yields similar results (frontiers changing by at most $1.8$
percentage points), indicating that the duplicate decisions have little effect on the reported frontier;
the formal guarantee itself rests on the stated one-decision-per-independent-patient sampling design rather
than on this empirical agreement.

\paragraph{Risk-level operating points.} Supplementary Table~\ref{stab:mondalpha} tabulates the
class-conditional deployment size and escalation fraction at $\alpha\in\{1,5,10\}\%$, giving the full
risk--resource trade-off for every model under the clinical default policy. The deployable
configuration that minimizes amortized size is chosen separately at each risk level, so the serving
configuration can differ across $\alpha$: at the strictest target ($\alpha=1\%$) several models fall
back from the NF4 $4$-bit scheme to the $8$-bit LLM.int8() scheme, or escalate in full. For example,
HuatuoGPT-o1-72B's $72.6\%$ amortized size at $\alpha=1\%$ is the $8$-bit scheme serving $54.7\%$ of
decisions with the remaining $45.3\%$ escalated to BF16, whereas Qwen2.5-72B remains on NF4 ($59.2\%$,
serving $54.4\%$ with $45.6\%$ escalated); a size of $100\%$ (for example Med42-70B, whose LLM.int8()
variant does not certify) marks models with no certified operating point at that target.

\paragraph{Split-sample exactness check.} The reported certificate uses the full calibration set for
both the candidate grid and the Clopper--Pearson bounds, which maximizes statistical power but makes the
grid a function of the calibration data and omits the corrections for configuration selection and
within-patient dependence. Re-running the class-conditional certificate in the exact split-sample form
defined above (grid on the $250$-decision selection subset, Clopper--Pearson bounds on the $996$-decision
one-per-patient certification subset, and an outer $\delta/H$ Bonferroni across the $H=2$ deployable
configurations; Algorithm~\ref{alg:certa}, Proposition~\ref{prop:cert} and Remark~\ref{rem:config})
reproduces the reported frontier closely (Supplementary Table~\ref{stab:exact}):
the deployment size changes by a median of $1.0$ and at most $8.2$ percentage points of BF16 size across
the eleven models, the flagship operating points move by about one point (Qwen2.5-72B $25.8\%$ to
$26.7\%$, HuatuoGPT-o1-72B $27.3\%$ to $27.9\%$), and every model still selects the NF4 $4$-bit
configuration. The reported higher-power frontier is therefore not an artefact of the data-dependent grid,
the configuration search, or within-patient dependence. The exact procedure's per-class certificate,
auditable from its served counts, disagreement counts, thresholds and Clopper--Pearson bounds, is given
in Supplementary Table~\ref{stab:exactaudit}; every served class satisfies $U_{\mathrm{CP}}\le5\%$.

\begin{table}[h]\centering\small
\caption{\textbf{Split-sample exactness.} Class-conditional deployment frontier (\% BF16, $\alpha=5\%$)
under the reported full-calibration certificate (grid and Clopper--Pearson bounds on all $1{,}250$
calibration decisions) and under the exact certificate (grid on the $250$-decision selection subset,
Clopper--Pearson bounds on the $1{,}000$-decision certification subset reduced to one decision per patient,
$996$ decisions, and an outer
$\delta/H$ Bonferroni across the $H=2$ deployable configurations; subsets mutually patient-disjoint). The difference column is exact minus full computed on the unrounded sizes and then rounded, so it may differ by $0.1$ from the difference of the two displayed (rounded) columns.
Every model selects the same NF4 $4$-bit deployable configuration and remains certified under the exact
procedure.}
\label{stab:exact}
\setlength{\tabcolsep}{7pt}
\begin{tabular}{lrrrcc}
\toprule
Model & \makecell{Full-cal.\\(\% BF16)} & \makecell{Exact\\(\% BF16)} & \makecell{Diff.\\(pp)} & \makecell{Frontier\\config} & \makecell{Exact\\certified?}\\
\midrule
\tableSexact
\bottomrule
\end{tabular}
\end{table}

\begin{table}[h]\centering\small
\caption{\textbf{Exact split-sample per-class certificate at the frontier ($\alpha=5\%$).} The
auditable formal guarantee underlying Table~\ref{stab:exact}: for each model, the class-conditional
certificate computed by the exact split-sample procedure (candidate grid on the $250$-decision selection
subset; Clopper--Pearson bounds on the one-decision-per-patient certification subset; outer Bonferroni
allocation $\delta/(2GH)$ with $H=2$ deployable configurations). For each decision class: $n$, served
certification observations; $x$, disagreements among them; $\tau$, selected margin threshold;
Cov, certified test coverage; $U_{\mathrm{CP}}$, one-sided Clopper--Pearson upper bound on the class risk.
Every served class satisfies $U_{\mathrm{CP}}\le5\%$. \emph{esc} ($\tau=\infty$): reject class escalated
in full, so no compressed decisions are served and the class contributes no compression-induced
disagreement. Each class certificate is reconstructible from $n$, $x$ and $\tau$.}
\label{stab:exactaudit}
\setlength{\tabcolsep}{4pt}
\resizebox{\textwidth}{!}{%
\begin{tabular}{lrrrrr rrrrr}
\toprule
& \multicolumn{5}{c}{Accept ($\mathrm{A}$)} & \multicolumn{5}{c}{Reject ($\mathrm{R}$)}\\
\cmidrule(lr){2-6}\cmidrule(lr){7-11}
Model & $n_{\mathrm{A}}$ & $x_{\mathrm{A}}$ & $\tau_{\mathrm{A}}$ & Cov (\%) & $U_{\mathrm{CP}}$ (\%)
      & $n_{\mathrm{R}}$ & $x_{\mathrm{R}}$ & $\tau_{\mathrm{R}}$ & Cov (\%) & $U_{\mathrm{CP}}$ (\%)\\
\midrule
\tableSexactaudit
\bottomrule
\end{tabular}}
\end{table}

\begin{table}[h]\centering\small
\caption{\textbf{Confidence-level sensitivity (marginal and class-conditional).} Frontier size
(\% BF16) at $\alpha=5\%$ for three confidence levels, for the marginal certificate and the
class-conditional (Mondrian) policy. Both change only modestly across $\delta$.}
\label{stab:conf}
\setlength{\tabcolsep}{7pt}
\begin{tabular}{lrrrrrr}
\toprule
& \multicolumn{3}{c}{Marginal} & \multicolumn{3}{c}{Class-conditional}\\
\cmidrule(lr){2-4}\cmidrule(lr){5-7}
Model & \makecell{$\delta{=}.10$\\(90\%)} & \makecell{$\delta{=}.05$\\(95\%)} & \makecell{$\delta{=}.01$\\(99\%)}
& \makecell{$\delta{=}.10$\\(90\%)} & \makecell{$\delta{=}.05$\\(95\%)} & \makecell{$\delta{=}.01$\\(99\%)}\\
\midrule
\tableSconfidence
\bottomrule
\end{tabular}
\end{table}

\begin{table}[h]\centering\small
\caption{\textbf{Calibration-size sensitivity.} Marginal frontier range (\% BF16) at $\alpha=5\%$
across the eleven models, as the patient-disjoint calibration set is reduced from $1{,}250$ to
$250$ decisions; the certificate degrades gracefully and the strongest models retain the $4$-bit floor.}
\label{stab:calsize}
\setlength{\tabcolsep}{10pt}
\begin{tabular}{rrrr}
\toprule
$n_{\mathrm{cal}}$ & \makecell{min\\($4\times$ floor)} & median & \makecell{max\\(least compressible)}\\
\midrule
\tableScalsize
\bottomrule
\end{tabular}
\end{table}

\begin{table}[h]\centering\small
\caption{\textbf{Threshold-grid sensitivity.} Marginal frontier size (\% BF16) at $\alpha=5\%$,
$\delta=0.10$ as the Learn-then-Test grid size $G$ is varied. The first three value columns are the
minimum, median and maximum frontier across the eleven models; the last three are three representative
models. The frontier is nearly invariant to $G$.}
\label{stab:grid}
\setlength{\tabcolsep}{7pt}
\begin{tabular}{rrrrrrr}
\toprule
$G$ & \makecell{min\\($4\times$ floor)} & median & \makecell{max\\(least compr.)} & \makecell{Qwen2.5\\72B} & \makecell{Huatuo\\o1-72B} & \makecell{Llama\\3.3-70B}\\
\midrule
\tableSgrid
\bottomrule
\end{tabular}
\end{table}

\begin{table}[h]\centering\small
\caption{\textbf{Split stability.} Marginal frontier size (\% BF16) at $\alpha=5\%$, $\delta=0.10$
across $20$ independent patient-disjoint calibration/test splits: per-model median, interquartile range
and full range. The frontier is stable across resamples; the strongest models stay at the $4$-bit floor
on every split.}
\label{stab:split}
\setlength{\tabcolsep}{9pt}
\begin{tabular}{lrrrr}
\toprule
Model & median & \makecell{IQR\\(25--75\%)} & min & max\\
\midrule
\tableSsplit
\bottomrule
\end{tabular}
\end{table}

\begin{table}[h]\centering\small
\caption{\textbf{Risk-level operating points (class-conditional).} Class-conditional (Mondrian)
frontier size (\% BF16) and escalation fraction (\%) at three risk targets $\alpha\in\{1,5,10\}\%$
($\delta=0.10$, $\delta/(2G)$ allocation). Relaxing $\alpha$ lowers both the deployed size and the
escalation fraction toward the $4$-bit floor. The deployable configuration that minimizes amortized
size is selected independently at each risk level, so the serving configuration can change with
$\alpha$ (see text); a size of $100\%$ denotes no certified operating point at that target, for which
all decisions escalate to BF16.}
\label{stab:mondalpha}
\setlength{\tabcolsep}{6pt}
\begin{tabular}{lrrrrrr}
\toprule
& \multicolumn{2}{c}{$\alpha=1\%$} & \multicolumn{2}{c}{$\alpha=5\%$} & \multicolumn{2}{c}{$\alpha=10\%$}\\
\cmidrule(lr){2-3}\cmidrule(lr){4-5}\cmidrule(lr){6-7}
Model & size & esc. & size & esc. & size & esc.\\
\midrule
\tableSmondalpha
\bottomrule
\end{tabular}
\end{table}

\section{Distribution robustness: prevalence shift}
Patient-disjoint calibration removes patient leakage but does not, by itself, guarantee robustness to
a shift between the calibration and deployment decision distributions. Because the marginal
certificate controls disagreement only on average over the calibration prevalence, a deployment whose
accept/reject mix differs from calibration can drive the marginal certified-set disagreement above the
target, whereas the class-conditional certificate controls each decision class separately and
therefore remains below the target across the tested class-prevalence shifts, provided within-class exchangeability
holds. This tolerance is conditional: class-conditional certification can absorb a change in the
accept/reject mixture only when the within-class margin distributions themselves remain stable, and it does
not guarantee robustness to an arbitrary clinical distribution shift that alters those within-class
distributions. Supplementary Figure~\ref{sfig:prevalence} and Supplementary Table~\ref{stab:prev} report the
certified-set disagreement as the deployment reject-class prevalence is swept from $5\%$ to $50\%$,
holding the calibrated thresholds fixed, for all eleven models. For OpenBioLLM-70B, the marginal
certificate rises to $9.5\%$ at a balanced reject prevalence, well above the $5\%$ target, and for
Llama-3-70B and Qwen2.5-72B it drifts above $5\%$ at low reject prevalence. Taking the maximum over the
eleven models, the marginal certified-set disagreement exceeds the $5\%$ target at every deployment
prevalence ($5.7\%$ to $9.5\%$), whereas the class-conditional policy stays below the target for every
model at every prevalence (maximum over models $2.7\%$ to $3.9\%$). This supports adopting the
class-conditional policy as the default CERTA policy for the evaluated clinical verification setting.

\begin{table}[h]\centering\small
\caption{\textbf{Prevalence-shift stress test.} Certified-set disagreement (\%) versus deployment
reject-class prevalence, for the marginal and class-conditional (Mondrian) policies with fixed
calibrated thresholds. Values above the $\alpha=5\%$ target are the marginal certificate's
prevalence-induced violations; the class-conditional policy stays below target throughout.}
\label{stab:prev}
\setlength{\tabcolsep}{7pt}
\begin{tabular}{llrrrrrr}
\toprule
Model & Policy & \multicolumn{6}{c}{$\NO$ prevalence (\%)}\\
\cmidrule(lr){3-8}
 & & 50 & 40 & 30 & 20 & 10 & 5 \\
\midrule
\tableSprev
\bottomrule
\end{tabular}
\end{table}

\section{Deployment behaviour: escalated cases}
Supplementary Figure~\ref{sfig:escalated} contrasts the decisions the router serves with those it
escalates, at the NF4 frontier. Escalated inputs have an approximately $8.6$-fold smaller mean decision
margin ($0.93$ versus $7.95$ for certified inputs) and a much higher disagreement with the reference
(mean $45.3\%$ versus $2.9\%$). Escalation is therefore not arbitrary: the router preferentially refers low-margin decisions on which
disagreement with the reference is enriched, and serves the high-margin decisions on which the
compressed model is reliable.
 
The separation is distributional, not an artefact of a few extreme inputs. Panel~(a) plots the full
empirical cumulative distribution of the decision margin for escalated and certified inputs, pooled
over the eleven models. Over this pooled sample the empirical escalated CDF lies at or above the certified
CDF at every margin, with no crossings (empirical first-order stochastic dominance), so the escalated
group is shifted toward lower margins throughout the observed range: the median margin is $0.35$ for
escalated decisions against $5.12$ for certified ones, a roughly fifteen-fold gap at the median that
parallels the $8.6$-fold gap in means. This ordering establishes that one distribution is shifted below
the other; it does not imply that the two distributions are non-overlapping. Model- and class-specific
margin thresholds then determine which inputs are eligible for compressed service, serving decisions above
each selected threshold $\tau^\star$ (per model, and per class under the class-conditional policy) and
escalating the rest.
 
Panel~(b) shows that this partition tracks reference disagreement at the level of individual models, not
just in the pooled average. For every one of the eleven models, the certified-set disagreement falls at
or below the $5\%$ target (mean $2.9\%$), whereas the escalated-set disagreement is far higher (mean
$45.3\%$); the decision margin is thus an effective per-input proxy for the risk that compression flips
the decision, and escalation concentrates almost all of the compression-induced disagreement onto the
minority of low-margin cases routed to the BF16 reference, leaving the served majority reliable. We
emphasise that this analysis concerns disagreement with the BF16 reference, not clinical correctness:
whether the escalated disagreements are clinically consequential is the object of the future
clinician-adjudication study.
 
\section{Task-wise certification}
Supplementary Table~\ref{stab:taskwise} reports certified coverage and observed disagreement separately
for the two verifier tasks (named-entity and answer verification), averaged over models. The
routed decisions behave consistently across both tasks at every precision (for example NF4 serves
$94.6\%$ of named-entity and $91.5\%$ of answer decisions, with observed certified-set disagreement of
$2.3\%$ and $3.8\%$ respectively, both within the $5\%$ target), showing consistent empirical behaviour
across the two evaluated task types.
 
Two further patterns are visible in the table. First, the observed certified-set disagreement stays at
or below the $5\%$ target for both task types under every reported configuration (its largest value, over
the models with a non-empty certified set, is $4.1\%$, for answer verification under round-to-nearest
$4$-bit). Because the certificate is calibrated by margin and
decision class and is not separately stratified by task, these task-wise rates are descriptive evidence
of consistent behaviour across the two evaluated tasks rather than independent task-conditional
finite-sample guarantees. Second,
certifiability is governed by the quantizer rather than by the task. Round-to-nearest $4$-bit collapses
on both tasks (certified coverage $60.0\%$ for named-entity and $56.5\%$ for answer verification), the
deployable NF4 configuration recovers both ($94.6\%$ and $91.5\%$), and $8$-bit rounding leaves coverage
essentially complete throughout ($100\%$/$100\%$ at RTN-$8$); the pattern of certifiability therefore
follows the precision scheme, not the task. Answer verification is consistently the slightly harder of
the two to certify, with marginally lower coverage and marginally higher disagreement at every
memory-reducing configuration, plausibly because it requires integrating a supplied context and
question rather than judging a single candidate entity; the gap is small and never breaches the target.
That the same margin-based certificate behaves consistently across two quite different verification
tasks indicates the mechanism is not tuned to a single task format.

\begin{table}[h]\centering\small
\caption{\textbf{Task-wise certification.} Certified coverage and observed certified-set disagreement
(\%) for named-entity (NER) and answer (QA) verification at $\alpha=5\%$, $\delta=0.10$ (higher-power
operational analysis). Coverage is averaged over all eleven models, assigning zero coverage to any
configuration that certifies no decisions for a model. Certified-set disagreement is undefined for a
model that serves no decisions, so the disagreement columns are averaged only over the models with a
non-empty certified set: this is all eleven models for RTN-8, RTN-6 and NF4; nine models for INT8 (the two
whose \texttt{LLM.int8()} configurations failed to produce usable inference outputs, Llama-3-70B and
Med42-70B, serve nothing); and nine (NER) and ten (QA) models for RTN-4.}
\label{stab:taskwise}
\setlength{\tabcolsep}{9pt}
\begin{tabular}{lrrrr}
\toprule
Configuration & \makecell{NER\\coverage} & \makecell{NER\\disag.} & \makecell{QA\\coverage} & \makecell{QA\\disag.}\\
\midrule
\tableStaskwise
\bottomrule
\end{tabular}
\end{table}

\section{External human-gold validation}
Supplementary Table~\ref{stab:ner} gives the full disease named-entity F1 score against human annotations on
BC5CDR-Disease and NCBI-Disease, at BF16 and under round-to-nearest $8/6/4$-bit and the deployed NF4
$4$-bit configuration, together with precision and recall for the deployed NF4 configuration. These
benchmarks measure correctness against human labels, a different question from decision substitution;
the deployed NF4 configurations change the F1 score comparably to round-to-nearest $4$-bit quantization (mean
absolute change $0.045$ on BC5CDR and $0.045$ on NCBI), and the domain model that collapses under
round-to-nearest $4$-bit (MMed-Llama-3.1-70B, $-0.163$ and $-0.198$) recovers under the deployed NF4
configuration ($+0.078$ and $+0.091$).

\begin{table}[h]\centering\small
\caption{\textbf{External disease named-entity results.} F1 against human annotations at BF16 and
under each configuration; the last column is precision/recall for the deployed NF4 $4$-bit
configuration.}
\label{stab:ner}
\setlength{\tabcolsep}{6pt}
\begin{tabular}{lrrrrrc}
\toprule
Model & \makecell{F1\\BF16} & \makecell{F1\\RTN-8} & \makecell{F1\\RTN-6} & \makecell{F1\\RTN-4}
& \makecell{F1\\NF4} & \makecell{NF4\\P/R}\\
\midrule
\tableSner
\bottomrule
\end{tabular}
\end{table}

\paragraph{Paired human-gold correctness.} The comparison above is against the human gold in aggregate.
To test whether compression changes human-gold \emph{correctness} rather than only reference agreement,
we additionally compared each quantized configuration against the same model's BF16 run on the same
held-out documents, under the identical overlap scorer, and summarise the paired differences over the
eleven models with document-level bootstrap intervals (Supplementary Table~\ref{stab:nerpaired}). The
deployable NF4 configuration produces a small but statistically resolvable negative paired F1 change on
BC5CDR-Disease (paired $\Delta$F1 $-0.016$ $[-0.028,-0.003]$, interval excluding zero), while the paired
estimate on NCBI-Disease is centred on zero ($+0.000$ $[-0.012,+0.012]$). The paired NF4 changes are
smaller in magnitude than the round-to-nearest $4$-bit changes ($-0.022$ $[-0.038,-0.006]$ on
BC5CDR-Disease and $-0.029$ $[-0.043,-0.015]$ on NCBI-Disease) on both benchmarks, with the
BC5CDR-Disease difference comparatively modest and the NCBI-Disease estimate centred on zero. The
per-entity transitions (Supplementary Table~\ref{stab:nertrans}) show the same asymmetry: relative to
BF16, round-to-nearest $4$-bit both loses the most gold entities and introduces the most false
positives, while NF4 stays close to break-even. Compression's effect on human-gold correctness
therefore tracks the quantizer rather than the bit-width, mirroring the reference-agreement result;
these benchmarks concern disease named-entity recognition and do not label the primary clinical
verifier's individual decisions.

\begin{table}[h]\centering\small
\caption{\textbf{Paired human-gold differences on the external benchmarks.} Entity-level precision,
recall and F1 of each configuration minus the same model's BF16 run on the same documents, averaged over
the eleven models, with document-level $95\%$ bootstrap intervals ($1{,}000$ resamples); matching is the
overlap scorer used throughout. $|\Delta\mathrm{F1}|$ is the mean absolute per-model F1 change.}
\label{stab:nerpaired}
\setlength{\tabcolsep}{5pt}
\resizebox{\textwidth}{!}{%
\begin{tabular}{llcccc}
\toprule
Benchmark & Config & $\Delta$P & $\Delta$R & $\Delta$F1 & mean $|\Delta\mathrm{F1}|$\\
\midrule
BC5CDR-Disease & RTN-8 & $+0.004$ $[-0.001,+0.009]$ & $+0.005$ $[-0.001,+0.010]$ & $+0.004$ $[-0.000,+0.010]$ & $+0.005$ $[+0.005,+0.013]$\\
 & RTN-6 & $-0.000$ $[-0.009,+0.008]$ & $+0.019$ $[+0.011,+0.028]$ & $+0.016$ $[+0.009,+0.024]$ & $+0.019$ $[+0.015,+0.027]$\\
 & RTN-4 & $-0.040$ $[-0.053,-0.027]$ & $-0.017$ $[-0.036,+0.003]$ & $-0.022$ $[-0.038,-0.006]$ & $+0.057$ $[+0.046,+0.074]$\\
 & NF4 & $-0.006$ $[-0.017,+0.005]$ & $-0.011$ $[-0.025,+0.004]$ & $-0.016$ $[-0.028,-0.003]$ & $+0.045$ $[+0.037,+0.060]$\\
\midrule
NCBI-Disease & RTN-8 & $-0.002$ $[-0.010,+0.005]$ & $-0.000$ $[-0.007,+0.006]$ & $-0.001$ $[-0.008,+0.006]$ & $+0.011$ $[+0.007,+0.019]$\\
 & RTN-6 & $+0.003$ $[-0.009,+0.014]$ & $+0.002$ $[-0.006,+0.010]$ & $+0.002$ $[-0.007,+0.010]$ & $+0.018$ $[+0.012,+0.031]$\\
 & RTN-4 & $-0.046$ $[-0.063,-0.030]$ & $-0.021$ $[-0.036,-0.006]$ & $-0.029$ $[-0.043,-0.015]$ & $+0.044$ $[+0.037,+0.062]$\\
 & NF4 & $+0.029$ $[+0.014,+0.046]$ & $-0.006$ $[-0.019,+0.006]$ & $+0.000$ $[-0.012,+0.012]$ & $+0.045$ $[+0.035,+0.059]$\\
\bottomrule
\end{tabular}}
\end{table}

\begin{table}[h]\centering\small
\caption{\textbf{Gold-entity coverage transitions on the external benchmarks.} For each configuration
versus the same model's BF16 run, pooled over models and documents (recall side): gold entities both
recover (preserved), newly recovered, lost, or still missed; net coverage change with its $95\%$
bootstrap interval; and $\Delta$FP, the change in false positives (precision side). Overlap matching.}
\label{stab:nertrans}
\setlength{\tabcolsep}{4pt}
\begin{tabular}{llrrrrrc}
\toprule
Benchmark & Config & \makecell{Preserved\\covered} & \makecell{Newly\\covered} & \makecell{Lost\\coverage} & \makecell{Stayed\\missed} & \makecell{Net\\coverage} & $\Delta$FP\\
\midrule
BC5CDR-Disease & RTN-8 & 2964 & 55 & 34 & 1600 & $+21$ $[-2,+43]$ & $-16$ $[-35,+5]$\\
 & RTN-6 & 2938 & 138 & 60 & 1517 & $+78$ $[+40,+118]$ & $+12$ $[-23,+48]$\\
 & RTN-4 & 2564 & 363 & 434 & 1292 & $-71$ $[-163,+22]$ & $+206$ $[+136,+286]$\\
 & NF4 & 2733 & 217 & 265 & 1438 & $-48$ $[-107,+21]$ & $+82$ $[+33,+135]$\\
\midrule
NCBI-Disease & RTN-8 & 2763 & 73 & 71 & 3000 & $+2$ $[-41,+43]$ & $+0$ $[-24,+26]$\\
 & RTN-6 & 2727 & 123 & 107 & 2950 & $+16$ $[-34,+64]$ & $-8$ $[-62,+44]$\\
 & RTN-4 & 2349 & 362 & 485 & 2711 & $-123$ $[-220,-33]$ & $+88$ $[+16,+165]$\\
 & NF4 & 2516 & 302 & 318 & 2771 & $-16$ $[-92,+58]$ & $-62$ $[-137,+4]$\\
\bottomrule
\end{tabular}
\end{table}

\section{Illustrative worst-case amplification calculation}
We evaluated whether a worst-case deterministic propagation of the quantization-induced logit
perturbation could bound the change in decision margin without any distributional assumption. Let
$z_F(x)$ and $z_k(x)$ denote the answer-token logits produced by the BF16 reference and the $k$-bit
quantized model on input $x$, and write $\Delta z(x)=z_k(x)-z_F(x)$ for the quantization-induced
per-logit perturbation, with accept- and reject-token components $\Delta z_{\mathrm{A}}(x)=z_k(x,\mathrm{A})-z_F(x,\mathrm{A})$
and $\Delta z_{\mathrm{R}}(x)=z_k(x,\mathrm{R})-z_F(x,\mathrm{R})$, and $\lVert\Delta z(x)\rVert_\infty=\max_j|\Delta z_j(x)|$
its largest absolute entry. Writing $m_F(x)=|z_F(x,\mathrm{A})-z_F(x,\mathrm{R})|$ for the reference
decision margin, the reference decision is preserved whenever this margin exceeds the induced
perturbation of the accept--reject logit difference,
\[
m_F(x) > |\Delta z_{\mathrm{A}}(x)-\Delta z_{\mathrm{R}}(x)| .
\]
The change in the accept--reject logit difference is in turn controlled by the largest per-logit
perturbation,
\[
|\Delta z_{\mathrm{A}}(x)-\Delta z_{\mathrm{R}}(x)| \le 2\lVert \Delta z(x)\rVert_\infty ,
\]
so a conservative sufficient condition for preserving the reference decision is
\[
m_F(x) > 2\lVert \Delta z(x)\rVert_\infty .
\]
In a worst-case analysis this perturbation propagates layerwise and accumulates multiplicatively through
the residual and normalization structure as a product of per-block amplification factors. Let $c$ denote
a per-block worst-case amplification factor, an upper bound on the operator norm by which one transformer
block can magnify a perturbation; for attention and MLP blocks this factor generally exceeds one. Under
the illustrative assumption $c=1.5$ across $L\approx80$ blocks, the multiplicative amplification factor
$c^{L}$ exceeds $10^{14}$. This calculation is not a numerical bound on the final logit perturbation
$2\lVert\Delta z(x)\rVert_\infty$, because no model-specific bound on the initial perturbation and no
layerwise operator-norm bounds were established for the evaluated networks. It illustrates why a generic
worst-case amplification argument would not, by itself, provide a useful deterministic certificate for
these models: an amplification factor of this size indicates that any bound built from such a propagation
would be far too loose to be informative. This is the reason CERTA is built on the finite-sample,
distribution-level Learn-then-Test argument of the main text rather than on a deterministic per-input
certificate; the deterministic route is reported here only to document that it was considered and found
uninformative at this scale.

\section{Reproducibility}
Supplementary Table~\ref{stab:inventory} lists the eleven evaluated checkpoints with their architecture
family, adaptation type and parameter count. Two independent quantization routes were evaluated so that
the frontier reflects a property of the certificate rather than of any single implementation:
round-to-nearest quantization at $8/6/4$ bits (symmetric per-group scales over groups of $128$ output-channel
elements, each group's scale chosen by a mean-squared-error search over clipping ratios
$\{0.7,0.8,0.9,1.0\}$, BF16 storage) and the bitsandbytes library (NF4 $4$-bit with double
quantization and BF16 compute, and $8$-bit \texttt{LLM.int8()}). The certificate itself is implemented in
Python with NumPy and SciPy: the Clopper--Pearson upper bound is the regularized incomplete
beta-function inverse (\texttt{scipy.special.betaincinv}), and confidence intervals are
patient-clustered bootstraps for the clinical-verifier certificate and document-level bootstraps for the
external benchmarks, from $1{,}000$ resamples. The model generations themselves---the reference and
quantized decisions and the decision margins---were produced once and are released as recorded
per-input outputs; from these recorded outputs every CERTA certificate, table and figure in the main
text and this Supplement is regenerated with no further model inference. This covers the
per-configuration and per-class certificates (operational and exact split-sample), the robustness
analyses (confidence level, calibration size, threshold grid, split resampling, one-decision-per-patient
and prevalence shift), the escalation diagnostics and the task-wise summaries. The external
named-entity results are released as separate artifacts: the recorded per-model, per-dataset benchmark
outputs (\texttt{bench.csv}) regenerate the external-NER F1 table (Supplementary
Table~\ref{stab:ner}) and Figure~5 directly with the included overlap scorer, scored against the public
BC5CDR-Disease and NCBI-Disease gold corpora; the per-entity paired analysis
(Supplementary Tables~\ref{stab:nerpaired} and~\ref{stab:nertrans}) is produced by the released scoring
and document-level paired-bootstrap code from the per-document entity-prediction dumps, whose
regeneration requires re-running inference on the public benchmark documents with the model weights. A
driver script in the code release (\texttt{reproduce\_si.sh}; see Code availability for the repository)
recomputes the certificate-derived tables and figures, and the external-NER scoring and figures from the
released benchmark outputs.

\noindent The code release specifies its analysis dependencies (an \texttt{environment.yml} and a
\texttt{requirements.txt}) and the released version is tagged at the commit used to produce the reported
results. The upstream inference that produced the recorded logits used the Hugging Face Transformers
library and PyTorch with no network access; the exact compute environment (GPU model and count, and
CUDA and driver versions) is specified by the authors in the code release, and NF4 and \texttt{LLM.int8()}
quantization require a CUDA accelerator.

\begin{table}[h]\centering\small
\caption{\textbf{Model and inference configuration.} The eleven evaluated checkpoints with
architecture/model family (Q, Qwen; L, Llama; M, Mistral), adaptation (G, general-purpose;
D, domain-adapted), parameter count, released checkpoint identifier,$^{\dagger}$ and measured peak VRAM
at the BF16 reference and at the deployed NF4 $4$-bit configuration. Every model was
evaluated under the same quantizer set (round-to-nearest $8/6/4$-bit and the bitsandbytes NF4 $4$-bit
and \texttt{LLM.int8()} routes) with BF16 compute; the deployed frontier configuration is NF4 $4$-bit for all
eleven models.}
\label{stab:inventory}
\setlength{\tabcolsep}{5pt}
\resizebox{\textwidth}{!}{%
\begin{tabular}{llcrlcrr}
\toprule
Model & Family & Adapt. & \makecell{Params\\(B)} & HuggingFace checkpoint$^{\dagger}$ & \makecell{Compute\\dtype} & \makecell{BF16\\(GB)} & \makecell{NF4\\(GB)}\\
\midrule
Qwen2.5-72B & Qwen & G & 72 & \texttt{Qwen/Qwen2.5-72B-Instruct} & BF16 & 146.3 & 43.0\\
HuatuoGPT-o1-72B & Qwen & D & 72 & \texttt{FreedomIntelligence/HuatuoGPT-o1-72B} & BF16 & 146.1 & 42.7\\
Med42-70B & Llama & D & 70 & \texttt{m42-health/Llama3-Med42-70B} & BF16 & 141.9 & 41.2\\
OpenBioLLM-70B & Llama & D & 70 & \texttt{aaditya/Llama3-OpenBioLLM-70B} & BF16 & 141.7 & 41.0\\
Llama-3-70B & Llama & G & 70 & \texttt{meta-llama/Meta-Llama-3-70B-Instruct} & BF16 & 141.9 & 41.2\\
MMed-Llama-3.1-70B & Llama & D & 70 & \texttt{Henrychur/MMed-Llama3.1-70B} & BF16 & 141.9 & 41.2\\
Llama-3.3-70B & Llama & G & 70 & \texttt{meta-llama/Llama-3.3-70B-Instruct} & BF16 & 141.9 & 41.2\\
BioMistral-7B & Mistral & D & 7 & \texttt{BioMistral/BioMistral-7B} & BF16 & 14.8 & 4.6\\
Llama-3-8B & Llama & G & 8 & \texttt{meta-llama/Meta-Llama-3-8B-Instruct} & BF16 & 16.7 & 6.4\\
Mistral-7B & Mistral & G & 7 & \texttt{mistralai/Mistral-7B-Instruct-v0.1} & BF16 & 14.9 & 4.7\\
MMed-Llama-3-8B & Llama & D & 8 & \texttt{Henrychur/MMed-Llama-3-8B-EnIns} & BF16 & 16.7 & 8.7\\
\bottomrule
\end{tabular}}
\vspace{2pt}
\begin{minipage}{0.96\textwidth}\footnotesize
$^{\dagger}$Checkpoint identifiers give the released model families used; the authors confirm the exact
revisions in the code release. Peak VRAM is the measured allocation during inference; the same values
appear in main-text Table~1. The internal short key \texttt{llama3-70b-base}, used in
the released code and in the external named-entity logs, denotes \texttt{meta-llama/Meta-Llama-3-70B-Instruct}
(the general-purpose Llama-3 70B, as distinct from Llama-3.3-70B); it is not a pretrained base checkpoint.
The same eleven checkpoints were used for the clinical-verification and the external named-entity
experiments.
\end{minipage}
\end{table}

\clearpage
\section*{Supplementary Figures}

\begin{figure}[h]\centering
\includegraphics[width=0.86\textwidth]{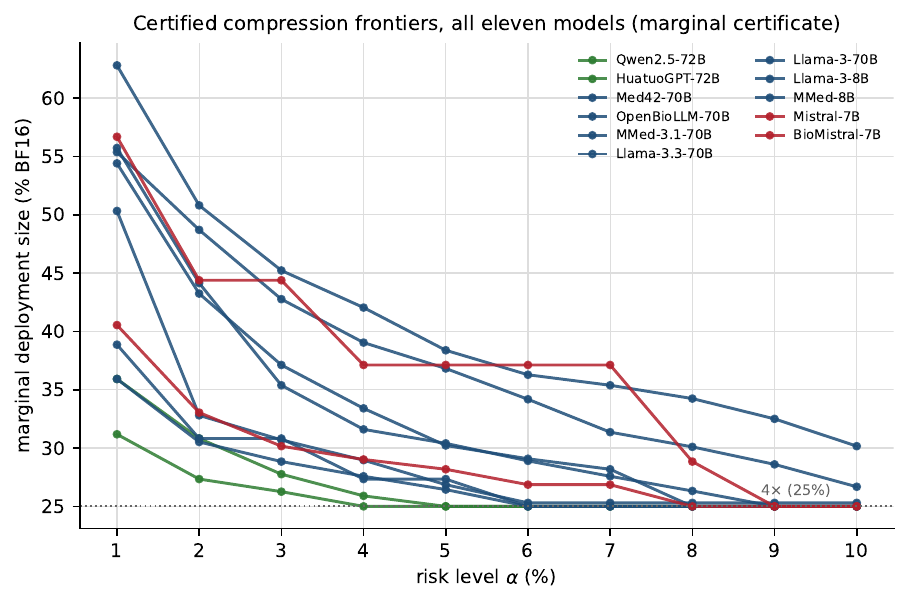}
\caption{\textbf{Certified compression frontiers for all eleven models under the marginal
certificate.} Marginal deployment size (\% BF16) as a function of the risk level $\alpha$; each curve
is one model, coloured by architecture family. The curves are monotonic and stable and settle toward
the $4$-bit floor ($25\%$, dotted) as $\alpha$ relaxes. This is the full version of the representative
subset in main-text Fig.~3a.}
\label{sfig:frontiers}
\end{figure}

\begin{figure}[h]\centering
\includegraphics[width=\textwidth]{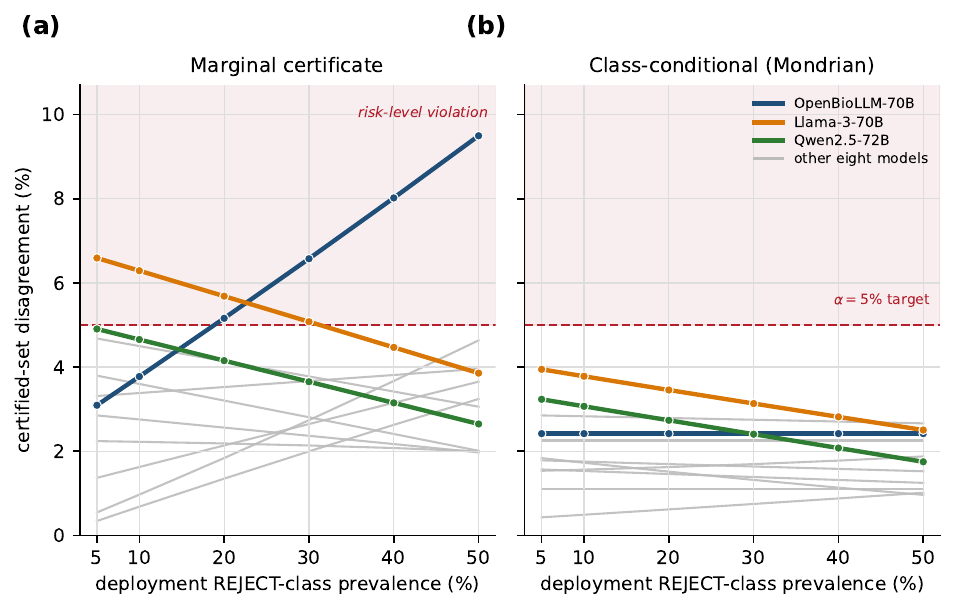}
\caption{\textbf{Prevalence-shift stress test, all eleven models.} Certified-set disagreement versus
deployment reject-class prevalence, with calibrated thresholds held fixed, for the
\textbf{(a)}, marginal and \textbf{(b)}, class-conditional (Mondrian) policies. All eleven models are
shown (three highlighted, the other eight in grey). Under the marginal certificate several models
enter the shaded risk-level violation zone above $\alpha=5\%$ (the maximum over models reaches
$9.5\%$); under the class-conditional policy every model stays below $\alpha$ at every prevalence
(maximum over models $3.9\%$; Supplementary Table~\ref{stab:prev}). The direction of the marginal
drift depends on which decision class carries more disagreement for each model.}
\label{sfig:prevalence}
\end{figure}

\begin{figure}[h]\centering
\includegraphics[width=\textwidth]{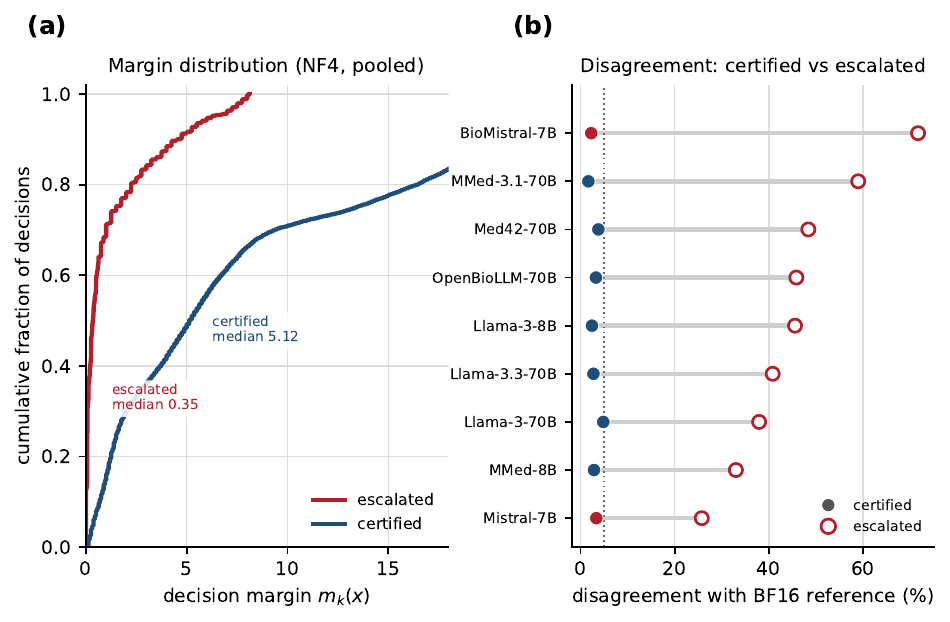}
\caption{\textbf{Escalated cases are enriched for low-margin, high-disagreement inputs
(NF4 $4$-bit, $\alpha=5\%$).}
\textbf{(a)}, Empirical cumulative distribution of the decision margin $m_k(x)$, pooled over the eleven
models, for escalated versus certified inputs; the distributions are strongly separated (median
margin $0.35$ escalated versus $5.12$ certified), not merely different in mean.
\textbf{(b)}, Observed disagreement with the BF16 reference for certified (filled) versus escalated
(open) inputs, per model: escalated inputs disagree with the reference far more often (mean $45.3\%$
versus $2.9\%$; dotted line, $5\%$). Escalation therefore selects the decisions on which the
compressed model is most likely to disagree with the BF16 reference; whether those disagreements are clinically consequential is the
object of the future clinician-adjudication study.}
\label{sfig:escalated}
\end{figure}

\begin{figure}[h]\centering
\includegraphics[width=\textwidth]{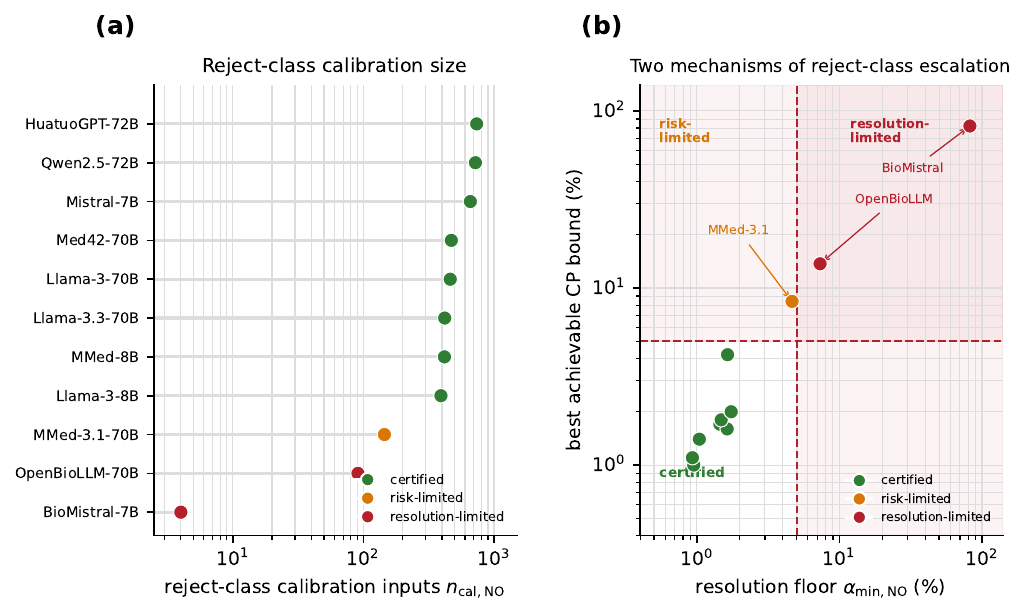}
\caption{\textbf{Two mechanisms of reject-class escalation.}
\textbf{(a)}, Number of $\NO$-class calibration inputs per model (log scale), coloured by
escalation mechanism: certified (green), risk-limited (orange), resolution-limited (red).
Acceptance-skewed models have very few reject-class calibration inputs (BioMistral-7B, $4$;
OpenBioLLM-70B, $91$).
\textbf{(b)}, Reject-class resolution floor $\alpha_{\min,\mathrm{R}}=1-(\delta/2G)^{1/n}$ (horizontal axis)
against the smallest attainable Clopper--Pearson upper bound on the reject-class risk (vertical axis),
both on log scales; dashed lines mark the $\alpha=5\%$ target. A class is certified only when both
coordinates fall below $\alpha$ (lower-left corner). Resolution-limited classes (BioMistral-7B,
OpenBioLLM-70B) fail the floor condition; the risk-limited class (MMed-Llama-3.1-70B) clears the floor
at $4.7\%$ but not the attainable bound at $8.4\%$. The floor is a necessary, not a sufficient,
certification condition.}
\label{sfig:classres}
\end{figure}

\begin{figure}[h]\centering
\includegraphics[width=\textwidth]{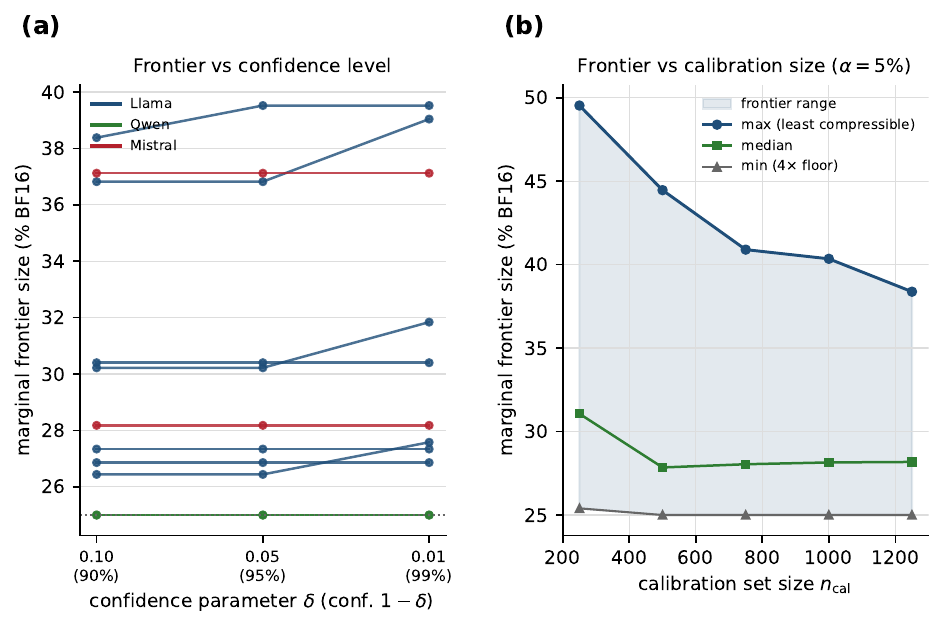}
\caption{\textbf{Statistical robustness.}
\textbf{(a)}, Marginal frontier size (\% BF16) at $\alpha=5\%$ for confidence levels
$\delta\in\{0.10,0.05,0.01\}$ (confidence $90/95/99\%$); each line is one model. The frontier
changes only modestly with the confidence level.
\textbf{(b)}, Frontier range (minimum, median, maximum over models) at $\alpha=5\%$ as a function of the
calibration set size $n_{\mathrm{cal}}$; the certificate degrades gracefully as calibration data are
reduced and tightens as they increase, while the strongest models retain the $4$-bit floor.}
\label{sfig:robust}
\end{figure}

\begin{figure}[h]\centering
\includegraphics[width=\textwidth]{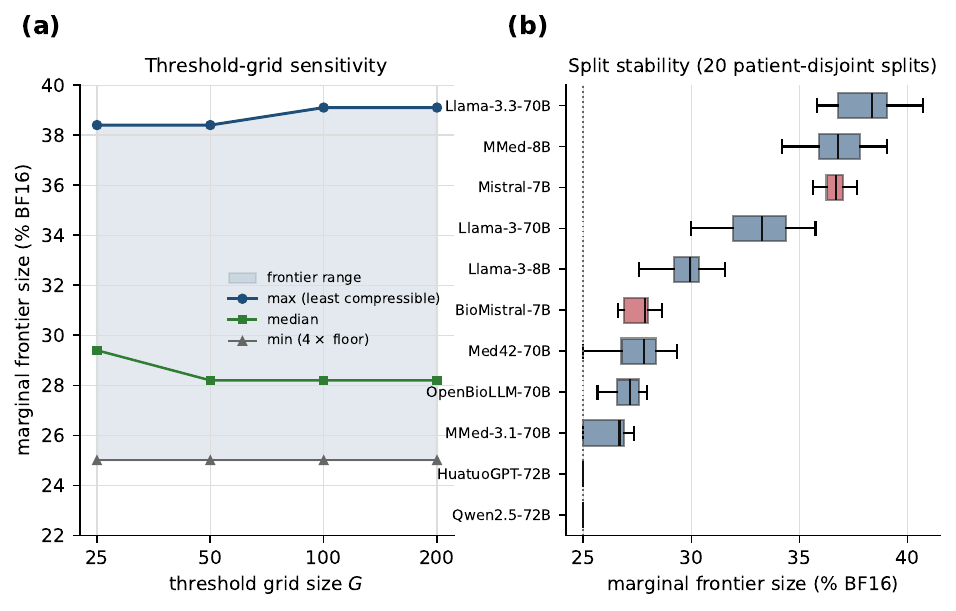}
\caption{\textbf{Design-choice robustness of the marginal frontier.}
\textbf{(a)}, Marginal frontier size (\% BF16) at $\alpha=5\%$ as the Learn-then-Test grid size $G$ is
varied over $\{25,50,100,200\}$: minimum ($4\times$ floor), median and maximum across the eleven
models. The envelope is nearly flat, so the frontier is insensitive to the grid resolution.
\textbf{(b)}, Distribution of the marginal frontier over $20$ independent patient-disjoint splits, one
boxplot per model (boxes coloured by architecture family, whiskers the full range; dotted line, the
$25\%$ $4$-bit floor). The strongest models sit at the floor on every split and no model's
interquartile width exceeds $2.5$ percentage points, confirming that the $25$--$40\%$ envelope is not
an artefact of the particular split used in the main text.}
\label{sfig:gridsplit}
\end{figure}

\section{Terminology and methodological rationale}
This note states, for each statistical, deployment and quantization term used in the main text and this
Supplementary Information, what it is and why it is the appropriate choice for a decision-level clinical
certificate. Symbols are defined in Table~\ref{stab:notation}, and the primary references for the named
methods are given in the main-text Methods.

\paragraph{BF16 reference.} The native $16$-bit (bfloat16) realization of each checkpoint, used as the
standard the quantized model must match. We adopt it because it is the model's own released precision, so any
disagreement is attributable to compression alone rather than to a change of model; all results are reported
relative to it.

\paragraph{Decision margin.} The absolute difference between the accept- and reject-answer logits,
$m=|z_{\mathrm{A}}-z_{\mathrm{R}}|$. It is a single confidence score the model already produces at no extra
inference cost, and it is the score CERTA thresholds on to determine which decisions are eligible to be
served by the quantized model under the specified reference-disagreement risk.

\paragraph{Conditional disagreement risk $\Risk(\tau)$.} The probability that the quantized and reference
decisions differ, conditioned on the margin exceeding $\tau$, i.e.\ evaluated only on served decisions. This
is the quantity the certificate bounds, because compression-induced disagreement is not incurred in the
final routed decision when that input is escalated to the BF16 reference.

\paragraph{Learn-then-Test.} A distribution-free calibration framework~\cite{angelopoulos2025learn} that selects a threshold on held-out
data with a finite-sample risk guarantee. We use it because it provides a genuine guarantee without
retraining and without parametric assumptions beyond patient-level iid sampling.

\paragraph{Binomial model.} Under the patient-level iid assumption and for a threshold fixed independently of
the certification sample, the number of disagreements among $n_\tau$ served inputs is
$\mathrm{Binomial}(n_\tau,\Risk(\tau))$. This exact model is what makes a finite-sample confidence bound
possible from counts alone.

\paragraph{Clopper--Pearson upper bound $U_{\mathrm{CP}}$.} The exact one-sided upper confidence limit~\cite{clopper1934} for a
binomial proportion, written in closed form as a Beta quantile
$U_{\mathrm{CP}}=\mathrm{Beta}^{-1}_{\,1-\delta'}(x_\tau+1,\,n_\tau-x_\tau)$. We use it rather than a normal
(Wald) approximation because it never under-covers, remains valid at small samples, and gives a correct,
non-degenerate bound when zero disagreements are observed, the small-sample setting most relevant to certification~\cite{brown2001interval}. Its
cost is statistical conservatism, which can become substantial when calibration samples are small, and this
conservatism favours escalation rather than uncertified deployment.

\paragraph{Bonferroni correction.} Because CERTA tests $G$ candidate thresholds and deploys the smallest that
passes, it performs a multiple comparison; evaluating each bound at level $1-\delta/G$ and taking a union
bound keeps the guarantee valid simultaneously across the whole grid. Bonferroni~\cite{dunn1961bonferroni} is used because it is
assumption-free (valid under arbitrary dependence) and transparent; we use $G=50$, and the grid-sensitivity
analysis (Supplementary Table~\ref{stab:grid}) shows the resulting frontier is essentially unchanged across
$G=25$--$200$. The marginal certificate allocates $\delta/G$ across the threshold grid; the
class-conditional certificate allocates $\delta/(2G)$ across the two classes and their grids. For the exact
frontier, which additionally selects among $H$ deployable configurations, these become $\delta/(GH)$ and
$\delta/(2GH)$ respectively (the exact class-conditional allocation is $\delta/(2GH)=5\times10^{-4}$ at
$\delta=0.10$, $G=50$, $H=2$).

\paragraph{Patient-level iid sampling.} The assumption that decisions, grouped by patient, are independent and identically distributed
draws from the deployment distribution. It is the sampling assumption underlying the binomial/Clopper--Pearson
argument; in the exact certificate, the candidate threshold grid is additionally fixed independently of the
certification sample; because sixteen patients contribute two decisions, inference and bootstrap resampling are performed
at the patient level, and a one-decision-per-patient re-analysis confirms the conclusions.

\paragraph{Class-conditional (Mondrian) certification.} Running the certificate separately within the
predicted accept and reject classes (a Mondrian, class-conditional scheme~\cite{vovk2003mondrian}), each at its own threshold and confidence allocation $\delta/(2G)$. It is
used because a single marginal guarantee can conceal a minority class with substantially higher disagreement;
conditioning on the class exposes and controls this asymmetry.

\paragraph{Zero-disagreement resolution floor $\alpha_{\min,g}$.} In the higher-power operational
class-conditional analysis, the smallest risk certifiable for a class with no observed disagreements given
its calibration count is $1-(\delta/2G)^{1/n_{\mathrm{cal},g}}$. This is the operational floor; the exact
class-conditional frontier instead uses the configuration-corrected allocation $\delta/(2GH)$ and the
class-specific count $n_{\mathrm{cert},g}$ in its one-decision-per-patient certification subset, giving
$1-(\delta/(2GH))^{1/n_{\mathrm{cert},g}}$.
It distinguishes classes escalated for lack of calibration data (\emph{resolution-limited}) from those
escalated for genuine disagreement (\emph{risk-limited}), making escalation interpretable rather than
arbitrary.

\paragraph{Patient-clustered bootstrap.} Confidence intervals obtained by resampling patients rather than
individual decisions, so that within-patient dependence is respected. It quantifies uncertainty on the
empirical coverage; the formal risk control is the Clopper--Pearson bound, not the bootstrap. For the
external named-entity benchmarks, whose units are documents rather than patients, the corresponding
intervals resample documents.

\paragraph{Risk level $\alpha$.} The target upper bound on the served-region disagreement risk, reported
from $1\%$ to $10\%$. It is the single knob a deployer sets to trade resource savings against the
tolerated rate of disagreement with the reference.

\paragraph{Confidence parameter $\delta$.} The probability with which the finite-sample guarantee is
allowed to fail over the draw of the calibration sample; we use $\delta=0.10$ (confidence $1-\delta=90\%$).
It is distinct from $\alpha$: $\alpha$ bounds the risk, whereas $\delta$ bounds the chance that the
certificate's risk bound is itself violated.

\paragraph{Candidate threshold grid ($G$).} The finite set of margin thresholds Learn-then-Test searches,
of size $G=50$. A finite grid makes the multiple-comparison correction explicit (Bonferroni over $G$); the
grid-sensitivity analysis (Supplementary Table~\ref{stab:grid}) shows the frontier is stable across
$G=25$--$200$.

\paragraph{Certified coverage ($C$).} The fraction of held-out decisions whose margin clears the selected
threshold and are therefore served by the quantized model, $C=\Prob(m_k(X)\ge\tau)$. It is the primary
benefit measure: higher coverage means more decisions avoid the full-precision reference.

\paragraph{Escalation rate ($E$).} The complementary fraction routed to the BF16 reference, $E=1-C$. It is
the cost side of the trade-off, since escalated decisions require the reference to be reachable.

\paragraph{Certify-or-escalate.} The deployment rule by which each input is either served by the quantized
model (if its margin clears the certified threshold) or escalated to the BF16 reference (otherwise). It is
the mechanism that converts a population-level risk bound into a per-decision policy with a guarantee.

\paragraph{Operating point.} A specific deployed choice of quantizer, bit-width and threshold at a given
risk level. Sweeping the risk level traces a model's operating points into its frontier.

\paragraph{Service score (routing statistic).} The scalar used to decide eligibility for compressed
service, here the decision margin $m_k(x)$. CERTA can use any prespecified scalar service score, including
a learned substitution-risk score, provided the score and the candidate threshold grid are fixed
independently of the certification sample; a score learned on separate selection data can therefore be
incorporated without changing the finite-sample argument. The margin is used here because the model already
produces it at no extra cost.

\paragraph{Empirical certified-set agreement ($\widehat A$).} The observed agreement rate between the
quantized and reference decisions on served inputs, $\widehat A=\Prob(Q_k(X)=F(X)\mid m_k(X)\ge\tau)$,
measured on held-out data. It is reported as an out-of-sample diagnostic; the formal guarantee is the
Clopper--Pearson bound, not $\widehat A$.

\paragraph{Marginal certificate.} A single threshold calibrated over both decision classes pooled
together. It is the simplest frontier and the baseline against which the class-conditional (Mondrian)
certificate is compared.

\paragraph{Accept and reject classes ($\mathrm{A}$, $\mathrm{R}$).} The two predicted decision classes of
the clinical verifier (accept or reject the candidate). The class-conditional certificate controls
disagreement within each separately, because the reject class is typically rarer and harder to certify.

\paragraph{Resolution-limited vs risk-limited escalation.} Two reasons a class is escalated in full:
\emph{resolution-limited}, when too few calibration decisions make even zero observed disagreements
insufficient to certify the target (its floor $\alpha_{\min,g}$ exceeds $\alpha$); and
\emph{risk-limited}, when enough data exist but the best attainable bound still exceeds $\alpha$. The
distinction separates a data-quantity limitation from a genuine disagreement problem.

\paragraph{Patient-disjoint calibration and testing.} The $2{,}500$ decisions (from $2{,}484$ patients)
are split so that no patient appears in both the $1{,}250$-decision calibration set and the
$1{,}250$-decision test set, preventing information about a patient from leaking between calibration and
evaluation.

\paragraph{Selection and certification subsets.} Within calibration, the exact certificate uses an
independent $250$-decision \emph{selection} subset to build the threshold grid and a disjoint
$1{,}000$-decision \emph{certification} subset (reduced to $996$ one-per-patient decisions) to compute the
Clopper--Pearson bounds. Separating them keeps the grid independent of the bounds, which the finite-sample
argument requires.

\paragraph{One decision per patient.} The certification unit of the exact certificate: at most one decision
is retained per patient, so that the served counts are independent, as the binomial argument requires. The
higher-power operational analysis instead uses all calibration decisions.

\paragraph{Higher-power operational analysis.} The reported full-calibration version of the certificate,
which builds the grid and computes the bounds on the same $1{,}250$ calibration decisions. It yields
tighter frontiers, but its grid is data-dependent, so its per-operating-point bounds are diagnostics rather
than an independent finite-sample guarantee.

\paragraph{Exact split-sample certificate.} The formally guaranteed version: independent selection and
certification subsets, one decision per patient, and an outer Bonferroni correction across the $H$
deployable configurations. It supplies the finite-sample guarantee, and the operational frontier is
reported as a higher-power comparison.

\paragraph{Deployable configurations ($H$).} The memory-reducing quantization schemes eligible for the
frontier, NF4 $4$-bit and $8$-bit \texttt{LLM.int8()} ($H=2$); round-to-nearest retains BF16 storage and is
excluded. The exact certificate corrects for selection among these $H$ configurations.

\paragraph{Compression-induced disagreement risk.} The quantity CERTA certifies: the rate at which the
quantized model's decision differs from the BF16 reference on served inputs. It is deliberately distinct
from \emph{clinical-correctness risk} (whether the reference decision is itself correct) and from
\emph{task-performance change} on the external benchmarks; the certificate bounds only the first.

\paragraph{Amortized size.} The coverage-weighted, BF16-equivalent deployment cost: $k$ bits for the
served fraction $C$ and $16$ bits for the escalated fraction give a bit-equivalent size
$B_{\mathrm{bits}}=16-(16-k)\,C$, reported throughout as a percentage of the BF16 reference,
\begin{equation*}
B_{\mathrm{amortised}}=100\,\frac{B_{\mathrm{bits}}}{16}
=100\left[1-\left(1-\tfrac{k}{16}\right)C\right]\quad(\%\ \text{of BF16}).
\end{equation*}
For example, a $4$-bit configuration serving every decision ($k=4$, $C=1$) has
$B_{\mathrm{amortised}}=25\%$. It is used so that a compressed model with occasional fallback can be
compared fairly against an uncompressed model; it is a cost-accounting quantity, not physical memory,
since escalated decisions still require a reachable reference.

\paragraph{Certified compression frontier.} For each model and target risk level $\alpha$, the
lowest-cost deployable configuration (bit-width and quantizer) whose certificate is admissible at that
$\alpha$; traced across $\alpha$ these operating points form the model's certified compression frontier,
the risk--resource trade-off reported throughout. A ``frontier point'' is one such selected
configuration.

\paragraph{Four-bit floor ($4\times$, $25\%$).} The smallest amortized size attainable by a deployable
$4$-bit configuration serving every decision, $k=4$ at $C=1$, i.e.\ $4/16=25\%$ of the BF16 nominal size (a
fourfold reduction). It is the lower limit a frontier can reach once escalation falls to zero, and the
dotted reference line in the risk--resource figures.

\paragraph{Round-to-nearest (RTN).} Nearest-level rounding kept at BF16 storage, hence not memory-reducing.
It is included as a numerical-precision baseline that isolates the effect of lower precision from the effect
of a memory-reducing kernel.

\paragraph{NF4.} The bitsandbytes $4$-bit NormalFloat format with double quantization, whose levels suit the
approximately Gaussian weight distribution and which genuinely reduces memory. It is the deployable $4$-bit
route and the configuration every model's frontier selects.

\paragraph{\texttt{LLM.int8()}.} A memory-reducing $8$-bit scheme with outlier handling. It is included as
the deployable $8$-bit counterpart to RTN-$8$ and, by failing on two BF16-native $70$B checkpoints,
demonstrates that numerical $8$-bit robustness and deployable $8$-bit memory reduction are distinct
properties.

\paragraph{Quantizer.} The scheme that maps BF16 weights to lower precision (round-to-nearest, NF4 or
\texttt{LLM.int8()}). The frontier is defined over quantizers, and two independent routes are evaluated, so
that the result reflects a property of the certificate rather than of any single implementation.

\paragraph{Nominal bit-width ($k$).} The configured per-weight precision of a quantizer ($4$ for NF4, $8$
for \texttt{LLM.int8()}). It enters the amortized-size accounting as a nominal figure and is not a
measurement of physical memory.

\paragraph{Peak VRAM.} The measured maximum GPU memory allocated during the forward pass, summed over the
devices a configuration requires. It is reported separately from amortized size because physical memory
also includes quantization metadata, activations and runtime allocations, and is therefore the quantity a
deployer provisions hardware against.

\paragraph{Overlap match (entity level).} The external-benchmark scoring rule by which a predicted disease
entity counts as a true positive if it overlaps a gold entity after whitespace and case normalization, and
a gold entity counts as recalled if some prediction overlaps it. It is used so that near-boundary
span differences are not penalized as outright errors.

\paragraph{Paired per-entity correctness and coverage transitions.} The document-level paired analysis that
compares each configuration against the same model's BF16 run on the same documents, reporting per-entity
transitions (gold entities preserved, newly covered, lost, or still missed) with document-level bootstrap
intervals. It isolates the compression effect on human-gold correctness from between-model variation.

\paragraph{Prevalence shift.} A stress test that reweights the deployment accept/reject mix away from the
calibration prevalence while holding the calibrated thresholds fixed. It probes whether the marginal
certificate's average guarantee survives a change in class balance, which the class-conditional certificate
is designed to withstand.

\bibliographystyle{unsrt}
\bibliography{sn-bibliography}